\documentclass[journal]{IEEEtran}
\newcommand{\ignore}[1]{}
\usepackage[utf8]{inputenc}
\usepackage[T1]{fontenc}
\usepackage{graphicx}
\usepackage{amsmath}
\usepackage{amssymb}
\usepackage{booktabs}
\usepackage{times}
\usepackage{epsfig}
\usepackage{cite}

\usepackage{algorithm}  
\usepackage{algpseudocode}
\usepackage{xcolor}
\usepackage{pifont}
\usepackage{multirow}
\usepackage{bbm}
\usepackage{mathtools}

\usepackage[pagebackref=true,breaklinks=true,colorlinks,bookmarks=false,linkcolor=black,anchorcolor=black,citecolor=black]{hyperref}
\usepackage[utf8]{inputenc} 
\usepackage[T1]{fontenc}    
\usepackage{hyperref}       
\usepackage{url}            
\usepackage{booktabs}       
\usepackage{amsfonts}       
\usepackage{nicefrac}       
\usepackage{microtype}      
\usepackage{xcolor}         

\usepackage{graphicx}
\usepackage{booktabs}

\usepackage{colortbl} 
\usepackage{pifont}
\usepackage{bbding}
\usepackage{fontawesome}
\usepackage{amsmath}
\usepackage{multirow}

\usepackage{xspace}
\newcommand{\ie}{{\emph{i.e.}}}
\newcommand{\eg}{{\emph{e.g.}}}

\newcommand{\corrops}{visual and textual correlation\xspace}
\newcommand{\Corrops}{Visual and Textual Correlation\xspace}
\newcommand{\CorrOps}{VTC\xspace}
\newcommand{\corragg}{multi-scale correlation aggregation unit\xspace}
\newcommand{\Corragg}{Multi-Scale Correlation Aggregation Unit\xspace}

\newcommand{\hscu}{high-level spatial correction unit\xspace}
\newcommand{\Hscu}{High-level Spatial Correction Unit\xspace}
\newcommand{\HSCU}{HSCU\xspace}
\newcommand{\eim}{embedding interactive unit\xspace}
\newcommand{\Eim}{Embedding Interactive Unit\xspace}
\newcommand{\EIM}{EIU\xspace}

\newcommand{\decoder}{cross-modal decoder\xspace}
\newcommand{\versus}{\textit{vs.}\xspace}

\begin{document}
\title{Beyond Mask: Rethinking Guidance Types in Few-shot Segmentation}
\author{Shijie~Chang,
        Youwei~Pang,
        Xiaoqi~Zhao,
        Lihe~Zhang,
        Huchuan~Lu
\thanks{
S. Chang, Y. Pang, X. Zhao, L. Zhang, and H. Lu are with Dalian University of Technology, China. (Email: csj@mail.dlut.edu.cn; lartpang@mail.dlut.edu.cn; zxq@mail.dlut.edu.cn; zhanglihe@dlut.edu.cn; lhchuan@dlut.edu.cn).
}}
\maketitle
\markboth{}{}

\begin{abstract}
Existing few-shot segmentation (FSS) methods mainly focus on prototype feature generation and the query-support matching mechanism. As a crucial prompt for generating prototype features, the pair of image-mask types in the support set has become the default setting. 
However, various types such as image, text, box, and mask all can provide valuable information regarding the objects in context, class, localization, and shape appearance. Existing work focuses on specific combinations of guidance, leading FSS into different research branches.
Rethinking guidance types in FSS is expected to explore the efficient joint representation of the coupling between the support set and query set, giving rise to research trends in the weakly or strongly annotated guidance to meet the customized requirements of practical users. 
In this work, we provide the generalized FSS with seven guidance paradigms and develop a universal vision-language framework (UniFSS) to integrate prompts from text, mask, box, and image. 
Leveraging the advantages of large-scale pre-training vision-language models in textual and visual embeddings, UniFSS proposes high-level spatial correction and embedding interactive units to overcome the semantic ambiguity drawbacks typically encountered by pure visual matching methods when facing intra-class appearance diversities. 
Extensive experiments show that 
UniFSS significantly outperforms the state-of-the-art methods.
Notably, the weakly annotated class-aware box paradigm even surpasses the finely annotated mask paradigm. 

\end{abstract}
\IEEEpeerreviewmaketitle


\section{Introduction}
Large-scale pixel-wise annotated datasets make significant progress in fully-supervised semantic segmentation. However, the labor cost of acquiring a large number of labeled datasets is very expensive. It takes more than an hour~\cite{cordts2016cityscapes,sakaridis2021acdc} to obtain pixel-level annotation of an image, making it difficult to add new categories. To address this challenge, the few-shot segmentation (FSS) task has been proposed. It aims to accurately delineate the corresponding object with an unseen category in the query image, utilizing merely a handful of reference exemplars as guidance. 

\begin{figure}[t]
  \includegraphics[width=\linewidth]{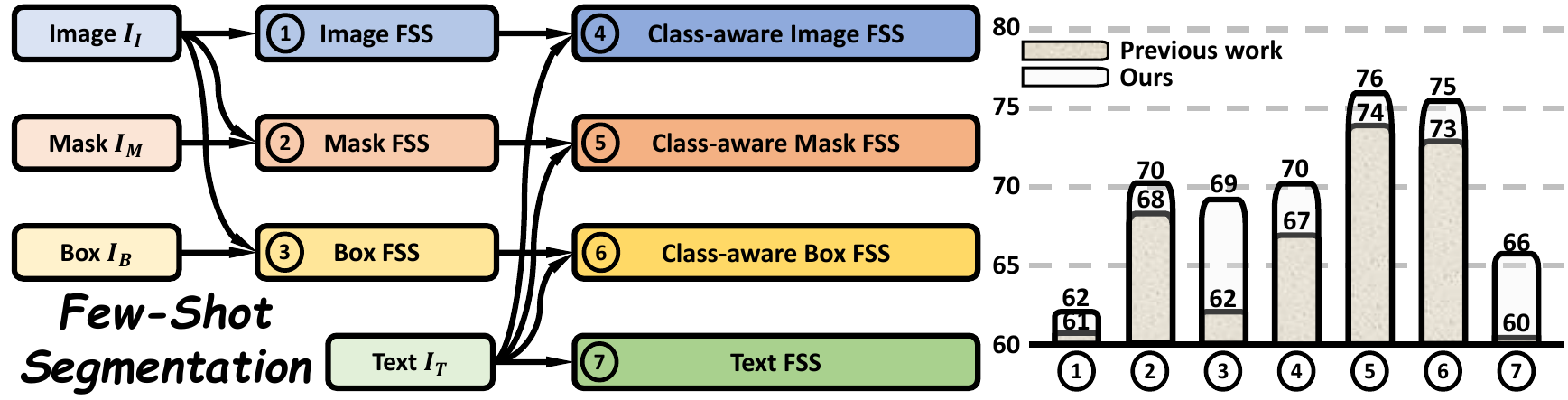}
  \centering
  \caption{All task patterns of few-shot segmentation corresponding to different combinations between image $I$, mask $M$, box $B$, and text $T$, and their mIoU performance comparison on PASCAL-$5^i$~\cite{shaban2017one}.
  Results of previous works are from:
  \cite{min2021hypercorrelation} for \ding{172},
  \cite{moon2023msi} for \ding{173}
  \cite{shi2022dense} for \ding{174},
  \cite{shuai2023pgmanet} for \{\ding{175}, \ding{176}, \ding{177}\},
  and \cite{liu2023delving} for \ding{178}. The performance results are rounded for easier representation.
  }
\label{fig:Figure1}
\end{figure}

According to the types of guidance information in the support set, we summarize seven patterns for few-shot segmentation, as shown in Fig.~\ref{fig:Figure1}. 
They are: \ding{172} Image FSS, \ding{173} Mask FSS, \ding{174} Box FSS, \ding{175} Class-aware Image FSS, \ding{176} Class-aware Mask FSS, \ding{177} Class-aware Box FSS, and \ding{178} Text FSS.
Existing works lack a comprehensive analysis of all task patterns, most methods primarily following the setting of pattern \ding{173} in~\cite{min2021hypercorrelation,tian2020prior,hong2022cost,peng2023hierarchical,fan2022ssp,shi2022dense}, 
and fewer studies focusing on patterns \ding{172} in~\cite{min2021hypercorrelation}, \ding{174} in~\cite{shi2022dense}, \{\ding{175}, \ding{176}, \ding{177}\} in~\cite{shuai2023pgmanet} and \ding{178} in~\cite{liu2023delving}. 
 Actually, the different task patterns are interconnected. For instance, \{\ding{173}, \ding{174}\} introduce dense annotations to \ding{172}, while \{\ding{175}, \ding{176}, \ding{177}\} introduce textual information to \{\ding{172}, \ding{173}, \ding{174}\}.
As the reference clues drive the segmentation of query target in FSS, the combination and balance of multi-granularity  guidance information from the support set is very important.
By rethinking multiple guidance clues, to establish an efficient semantic and appearance matching mechanism and promote FSS in a multi-granularity universal framework, is a crucial step towards general artificial intelligence (AGI).

Prototypical learning~\cite{yang2021mining, zhang2021prototypical, li2021adaptive, fan2022ssp, zhang2022mask, zhang2021self, liu2022dynamic, liu2022learning, zhang2022feature, okazawa2022interclass} and dense correlation learning~\cite{min2021hypercorrelation, hong2022cost, shi2022dense, moon2023msi, peng2023hierarchical, xiong2022doubly} are two popular research directions in FSS.
Unfortunately, these methods struggle to overcome the challenge of prompt ambiguity due to intra-class appearance diversities appearing in the limited number of support images.
Although the self-support matching strategy~\cite{fan2022ssp, wang2023focus} leverages query prototypes to match query itself features, the low-quality initial coarse segmentation maps further introduce interference from background information. MIANet~\cite{yang2023mianet} adopts word embeddings to obtain general high-level semantic information.
However, the independent textual embedding space cannot align well with the visual embedding space pre-trained on the ImageNet classification task.
In recent years, large-scale vision-language pre-training models such as CLIP~\cite{radford2021clip} have become popular in many cross-modal fields~\cite{shi2022proposalclip,du2022learning,zhou2022maskclip} because they can provide well-aligned textual and visual embeddings. 
Most recently, PGMANet~\cite{shuai2023pgmanet} employs CLIP as feature extractor, while it only treats textual embeddings as an unlearnable prior, neglecting the potential capability between textual and visual information. How to utilize the visual and textual embeddings produced by CLIP is key to solving intra-class diversities in FSS task.
Moreover, previous works design models for limited task patterns. 
However, the model's utilization of different guidance types exhibits similarities. An advanced model's ability to leverage guidance information should be reflected across all 7 task patterns.
A unified architecture capable of completing 7 task patterns should be explored.

To this end, we develop the universal vision-language few-shot segmentation framework (UniFSS) for adapting the multi-granularity guidance clues, including image, text, box, and mask.
Firstly, we compute multi-scale visual-visual and visual-textual correlations based on well-aligned visual and textual embeddings extracted by the pre-trained CLIP model.
To address the spatial information loss caused by large-stride downsampling in high-level features, we introduce the \hscu (\HSCU) to enhance the localization capability of high-level correlation features.
Secondly, we represent both visual-visual and visual-textual correlations in a unified 4D form and accomplish multi-scale correlation aggregation.
Lastly, we design the  \eim (\EIM) to fuse query and guidance (textual or masked average support image embeddings) features via Transformer style and combine with prepared 4D correlations to achieve cross-modal decoding.
We hope that our work can advance the development of FSS and promote the unification of FSS tasks.

Our contributions are summarized as follows:
\begin{itemize}
    \item We rethink the guidance types in FSS. It is the first systematic summary and assessment of different types of support set forms from a macroscopic perspective. 
    We attempt to integrate the different task branches of FSS.
    \item We design a universal vision-language framework, UniFSS, which is compatible with both visual-visual and vision-textual correlations, integrating the complementary strengths of appearance, shape, local and global context to overcome the challenges from intra-class diversities. As a universal architecture, UniFSS can accomplish 7 task patterns without modifying the model.
    \item Experimental results reveal the prompt capabilities of different types and combinations of support sets for query images. UniFSS significantly surpasses the existing state-of-the-art algorithms across different task patterns on three popular FSS benchmarks PASCAL-5$^i$~\cite{shaban2017one}, COCO-20$^i$~\cite{lin2014microsoft}, and FSS-1000~\cite{li2020fss1000}. 
\end{itemize}

\section{Related work}
\subsection{Few-shot Segmentation.} 
Few-shot segmentation (FSS) is first introduced in OSLSM~\cite{shaban2017one}. Subsequent works have proposed various new techniques to improve FSS performance based on metric-based learning paradigm~\cite{dong2018few}.
Inspired by PrototypicalNet~\cite{snell2017prototypical}, prototypical learning methods propose novel approaches to extract single~\cite{yang2021mining,wang2019panet,fan2022ssp,zhang2022mask}, or multiple prototypes~\cite{zhang2021self,liu2022dynamic,liu2022learning,zhang2022feature,okazawa2022interclass} and then enhance the effectiveness of feature aggregation by these prototypes.
For instance, SSP~\cite{fan2022ssp} extracts the self-support prototype and adaptive self-support background prototype from the query to enhance the representation capability of the support prototype.
Benefiting from the success of attention mechanism~\cite{vaswani2017attention,liu2021swin} and visual correspondence~\cite{cho2021cats,li2020correspondence}, dense correspondence learning methods~\cite{min2021hypercorrelation,peng2023hierarchical,hong2022cost,shi2022dense,xiong2022doubly,wang2023rethinking} introduce new cross-attention or correlation aggregation approaches to achieve pixel-level aggregation.
For example, HDMNet~\cite{peng2023hierarchical} designs hierarchical dense correlation distillation to improve cross-attention, while VAT~\cite{hong2022cost} proposes a 4D Swin Transformer to enhance correlation aggregation.
Compared with detailed prototype learning and dense correlation technology, the guidance information of the support set fundamentally determines the matching form and information content between the query and support.

\subsection{Guidance Types of Support Sets.}
Different types of guidance information play crucial roles in both high-level and low-level aspects. Images can provide texture and color information, assisting the model in capturing finer details of the targets. Boxes highlight position and size information, enabling the model to more accurately locate objects and understand their relative relationships in the scene. Masks contain detailed shape information, aiding the model in more accurately understanding the boundaries of objects. Text can clarify the category and context of the objects.
Different combinations of guidance types can produce seven research patterns, as shown in Fig.~\ref{fig:Figure1}.
Most works~\cite{min2021hypercorrelation,tian2020prior,hong2022cost} are designed for the \ding{173} setting. Some models~\cite{min2021hypercorrelation,shi2022dense} are developed for the \{\ding{172},\ding{174}\} settings by removing masks or replacing masks with boxs. Textual information is introduced in ~\cite{siam2020weakly,zhang2022weakly,wang2022imr} to address \ding{176} by generating pseudo labels for support images. Several works~\cite{xu2021simple,li2022languagedriven,lueddecke22clipseg} utilize the CLIP~\cite{radford2021clip} model to specifically tackle the pattern \ding{178}. 
Recently, ~\cite{shuai2023pgmanet} integrates all class-aware tasks \{\ding{175},\ding{176},\ding{177},\ding{178}\} in a single model architecture and achieves impressive results on these tasks..
The above works lead FSS towards different task branches \{\ding{172}-\ding{178}\}.
In this paper, guidance types are discussed in detail in a universal framework for the first time.
We integrate 7 different task branches into a unified branch, demonstrating that an advanced model architecture can accomplish all task patterns.

\subsection{Vison-Language Models.}
Pre-trained vision-language models (VLMs)~\cite{jia2021scaling,cherti2023reproducible,Radford2021LearningTV,radford2021clip, dou2022coarse, zhou2022learning} which learn vision-language correspondence has demonstrated great effectiveness in many downstream vision tasks.
Various unsupervised and open-vocabulary downstream tasks can benefit from VLMs.
CLIP~\cite{radford2021clip} itself can be used directly for zero-shot image classification with remarkable results.
ProposalCLIP~\cite{shi2022proposalclip} exploits CLIP to predict proposals for a large variety of object categories without annotations.
DetPro~\cite{du2022learning} learns continuous prompt representations for open-vocabulary object detection based on VLMs.
CLIPseg~\cite{lueddecke22clipseg} proposes a lightweight transformer-based decoder to adapt CLIP~\cite{radford2021clip} to accomplish segmentation tasks.
ZegFormer~\cite{ding2022decoupling} and ZSseg~\cite{xu2021simple} proposed two-stage frameworks for open-vocabulary segmentation.
Specifically, they first learn to predict class-agnostic region proposals and feed them to CLIP for final predictions.
In this work, we fully explore the potential of CLIP for few-shot segmentation.
Based on vision-language model, we integrate class-aware FSS \{\ding{175},\ding{176},\ding{177},\ding{178}\} and \{\ding{172},\ding{173},\ding{174}\} into a universal architecture.

\begin{figure*}[t]
  \centering
  \includegraphics[width=0.9\linewidth]{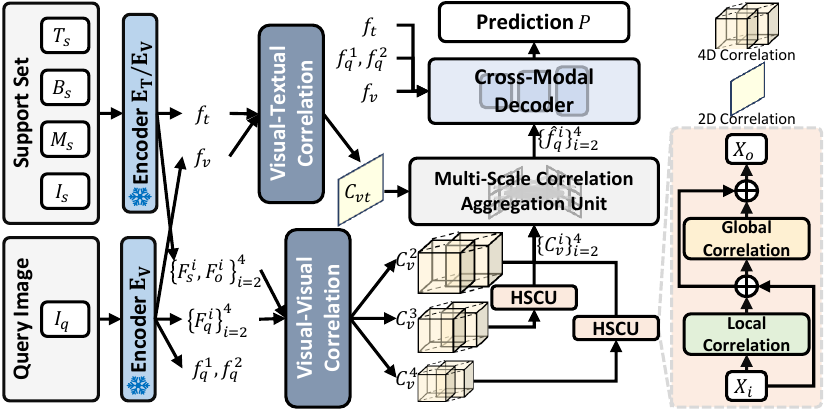}
  \caption{Overview of the proposed UniFSS framework. The proposed framework consists of four components, \ie, 1) \corrops (Sec. \ref{sec:corr_computation}), 2) \hscu (Sec. \ref{sec:hscu}), 3) \corragg (Sec. \ref{sec:ms_corragg}) and 4) \decoder (Sec. \ref{sec:decoder}).
  }
  \label{fig:pipeline}
\end{figure*}

\section{Methodology}

\label{sec:meth}


\subsection{Problem Formulation}
\label{sec:task}

In FSS, the dataset is divided into $\mathcal{D}_\mathrm{base}$ with category set $\mathcal{C}_\mathrm{base}$ for training and $\mathcal{D}_\mathrm{novel}$ with unseen category set $\mathcal{C}_\mathrm{novel}$ for testing.
$\mathcal{C}_\mathrm{base}$ and $\mathcal{C}_\mathrm{novel}$ are disjoint, \ie, $\mathcal{C}_\mathrm{base} \cap \mathcal{C}_\mathrm{novel} = \emptyset$.
FSS aims to train a model that can segment objects from unseen $\mathcal{C}_\mathrm{novel}$ in a query image given only a few reference exemplars with the same class.
Following~\cite{tian2020prior,min2021hypercorrelation,hong2022cost}, episodes are applied to both train set $\mathcal{D}_\mathrm{base}$ and test set $\mathcal{D}_\mathrm{novel}$.
Each episode consists of a query set $\mathcal{Q} = \{I_q, M_q\}$ and support set $\mathcal{S} = \{\mathcal{S}_k\}_{k=1}^K$ with the same category $c$ of interest in the $K$-shot setting.
$\mathcal{S}_k$ usually involves different types of data, including RGB image $I_s$, binary mask $M_s$, box $B_s$, and class text $T_s$ as shown in Fig.~\ref{fig:Figure1}.
During training, the model is trained to learn a mapping from $\mathcal{S}$ and $I_q$ to $M_q$ under the supervision of $M_q$.
Specifically, the model $f(\cdot,\theta)$ learns to predict a binary mask $P$ for $I_q$ with the guidance of $\mathcal{S}$ for a specific novel class $c \in \mathcal{C}_{novel}$, \ie,
$P = f(\mathcal{S}, I_q, \theta | c)$.
The trained model is then evaluated on the episodes sampled from $\mathcal{D}_\mathrm{novel}$ without further optimization.

\subsection{Overview of Our Method}
\label{sec:network}

Without loss of generality and simplicity, we introduce the proposed UniFSS framework in the context of class-aware image FSS under the 1-shot setting, \ie, $\mathcal{S} = \{I_s, M_s, T_s\}$.
And the adjustments for other FSS patterns are described in Sec. \ref{sec:task_patterns}.
The overall architecture is illustrated in  Fig.~\ref{fig:pipeline}.
Specifically, given $I_q$ and $\mathcal{S}$, we first get the object image $I_o$ by masking $I_s$ with the support mask $M_s$, \ie, $I_o = I_s \odot M_s$, where $\odot$ denotes the Hadamard product.
And, the frozen CLIP encoders $\mathbf{E_V}$ and $\mathbf{E_T}$ are utilized to extract stage-wise visual features $f_q^1, f_q^2$ and visual embedding $f_v$ of $I_q$, and textual embedding $f_t$ of $T_s$.
We compute the visual-textual correlation $C_{vt}$ between $f_v$ and $f_t$.
Besides, following MSI operation~\cite{moon2023msi}, the hierarchical layer-wise features $F_q = \{F_q^i\}_{i=2}^{4}$ of $I_q$, $F_o = \{F_o^i\}_{i=2}^{4}$ of $I_o$ and $F_s = \{F_s^i\}_{i=2}^{4}$ of $I_s$ are collected from the layers in the encoder stages $\{2,3,4\}$ to compute the multi-scale visual dense correlations $C_{v} = \{C^{i}_{v}\}_{i=2}^{4}$.
$C_{vt}$ and $C_{v}$ are fed into the \hscu and \corragg for the guidance refinement and correlation aggregation.
Finally, the accurate prediction $P$ is obtained by decoding the aggregated representation $\hat{F}_q = \{\hat{f}_q^i\}_{i=2}^{4}$, textual and visual embeddings $f_t$ and $f_v$, and low-level query features $f_q^1, f_q^2$ in the \decoder.

\begin{figure*}[t]
  \centering
  \includegraphics[width=\linewidth]{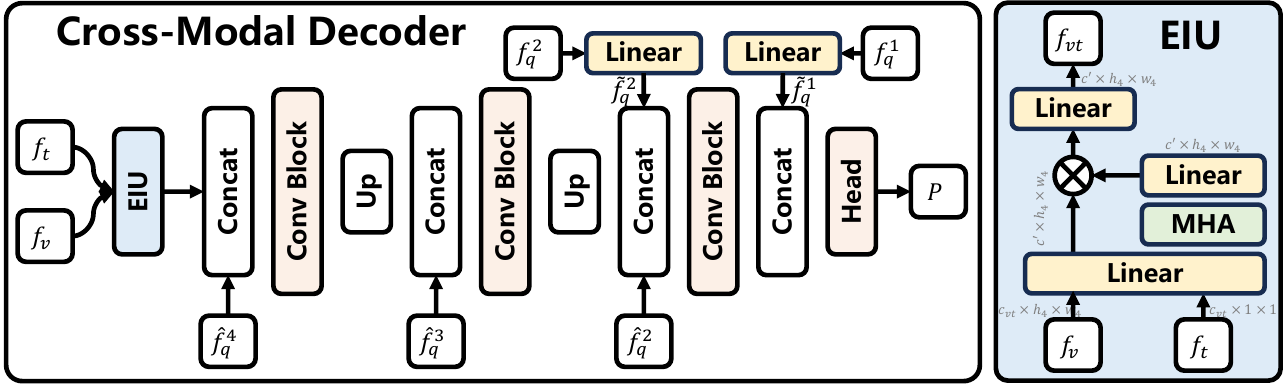}
  \caption{Details of our \decoder with the proposed \eim (\EIM).
  ``Up'': bilinear interpolation.
  ``$\otimes$'': Hadamard product.}
  \label{fig:decoder}
\end{figure*}

\subsection{\Corrops (\CorrOps)}
\label{sec:corr_computation}

\textbf{Vision and Text.}
Inspired by MaskCLIP~\cite{zhou2022maskclip}, we reformulate the value embedding layer and the last linear layer into two respective $1 \times 1$ convolutional layers to obtain the visual embedding $f_{v} \in \mathbb{R}^{c_{vt} \times h_4 \times w_4 }$ for the query input $I_q$.
And the textual embedding $f_{t} \in \mathbb{R}^{c_{vt} \times 1 \times 1}$ is extracted by the CLIP encoder $\mathbf{E_T}$ from the corresponding class text $T_s$.
The non-parametric cosine similarity is introduced to obtain the visual-textual dense correlations $C_{vt}$ between the deep embeddings $f_v$ from $I_q$ and $f_t$ from $T_s$.
This process is formalized as
\begin{equation}
  C_{vt} = \mathrm{ReLU}\left(
        \frac{\tilde{f}_v \cdot {\tilde{f}_t}^{\top}}
             {\left \| \tilde{f}_v \right \| \cdot \left \| \tilde{f}_t \right \|}
    \right),
  \label{eq:corr_vt}
\end{equation}
where
$\tilde{f}_v \in \mathbb{R}^{h_4 w_4 \times c_{vt}}$ and $\tilde{f}_t \in \mathbb{R}^{1 \times c_{vt}}$ are $f_{v}$ and $f_t$ after reshaping, respectively.
By \ref{eq:corr_vt}, we can obtain the visual-textual correlation map $C_{vt} \in \mathbb{R}^{h_4 w_4 \times 1}$.

\textbf{Vision and Vision.}
In the same way, we also calculate the cosine similarity to obtain the multi-scale visual dense correlations $C_{v} = \{C^{i}_{v}\}_{i=2}^{4}$ between the hierarchical layer-wise features $F_q = \{F_q^i\}_{i=2}^{4}$ of $I_q$, $F_o = \{F_o^i\}_{i=2}^{4}$ of $I_o$ and $F_s = \{F_s^i\}_{i=2}^{4}$ of $I_s$ as follows:
\begin{equation}
  C_{v}^{i,l} = \mathrm{ReLU}\left(\frac{\tilde{F}_q^{i,l} \cdot {\tilde{F}_s^{i,l\top}}}{\left \| \tilde{F}_q^{i,l} \right \| \cdot \left \| \tilde{F}_s^{i,l} \right \|}\right) \oplus \mathrm{ReLU}\left(\frac{\tilde{F}_q^{i,l} \cdot {\tilde{F}_o^{i,l\top}}}{\left \| \tilde{F}_q^{i,l} \right \| \cdot \left \| \tilde{F}_o^{i,l} \right \|}\right),
  \label{eq:corr_v}
\end{equation}
where $\oplus$ indicates the concatenation operation and $(i,l)$ indexes the feature map from layer $l$ in encoder stage $i$.
$\tilde{F}_q^{i,l} \in \mathbb{R}^{h_{i} w_{i} \times c_{i}}$, $\tilde{F}_t^{i,l} \in \mathbb{R}^{h_{i} w_{i} \times c_{i}}$ and $\tilde{F}_s^{i,l} \in \mathbb{R}^{h'_{i} w'_{i} \times c_{i}}$ are $F_q^{i,l}$, $F_o^{i,l}$ and $F_s^{i,l}$ after reshaping.
The layer-wise correlation map $C_{v}^{i,l} \in \mathbb{R}^{2 \times h_{i} w_{i} \times h'_{i} w'_{i}}$ provides more fine-grained mining of object similarity clues.
Following~\cite{min2021hypercorrelation,hong2022cost,moon2023msi}, the correction maps from the same stage are concatenated along channel dimension to form initial visual correction map set $C_{v} = \{C_v^{i}\}_{i=2}^4$, where $C_{v}^i \in \mathbb{R}^{L_i \times h_{i} \times w_{i} \times h'_{i} \times w'_{i}}$ and $L_i$ is the number of layers in stage $i$.

\subsection{\Hscu (\HSCU)}
\label{sec:hscu}

Existing methods~\cite{min2021hypercorrelation,hong2022cost, moon2023msi} feed the initial correlation maps directly into the correlation aggregation module.
However, they ignore the problem of inaccurate localization in high-level correlation maps calculated from large-stride downsampling feature maps.
Incorrect high-level correlation maps will misguide the correlation aggregation of low-level correlation maps in the following procedures.
Therefore, the \HSCU is proposed to refine the high-level correlations $C_v^4$ and $C_v^3$.
For simplicity, $C_{v}^4$ is used here to illustrate the process.

\textbf{Point-wise Local Correction.}
Specifically, the high-level correlation map $C_{v}^4 \in \mathbb{R}^{L_4 \times h_4 \times w_4 \times h'_4 \times w'_4}$ can be viewed as a feature map $X_i \in \mathbb{R}^{L_4 \times h'_4 w'_4 \times h_4 \times w_4}$ with  batch size of $L_4$,  spatial dimension of $h_4 \times w_4$, and channel dimension of $h'_4 w'_4$.
$X_i$ is locally corrected as follows:
\begin{equation}
  X' = X_i + \mathrm{DWConv}(X_i),
  \label{eq:dwconv}
\end{equation}
where $\mathrm{DWConv}$ denotes the depth-wise convolution layer.
Such a point-wise operation refines the correspondence between the local region of the query feature maps and specific points of the support feature maps.

\textbf{Global Information Correction.}
After point-wise local correction, a simple multi-layer perception (MLP) is used for global information correction as follows:
\begin{equation}
  X_o = X' + \mathrm{MLP}(\mathrm{LN}(X')),
  \label{eq:mlp}
\end{equation}
where $\mathrm{LN}$ denotes the layer normalization~\cite{ba2016ln}.
MLP operates on the spatial dimension $h'_4 w'_4$ of the support feature  to achieve global spatial correction of the support feature, calibrating the specific point-wise representation in the query feature map with the global support context.
Finally, the high-level correlation map $C_v^4$ is replaced by $X_o$ after adjusting the shape.

\subsection{\Corragg}
\label{sec:ms_corragg}

\textbf{Correlation Aggregation.}
In Sec. \ref{sec:corr_computation}, we obtain the 2D visual-textual correlation map $C_{vt}$ and further broadcast it as a 4D correlation map $C'_{vt} \in \mathbb{R}^{1 \times h_4 \times w_4 \times h'_4 \times w'_4}$.
$C_{vt}$ and $C'_{vt}$ show the activation of the query visual feature $f_v$ by the textual embedding $f_t$, which can well guide the visual correlation aggregation.
There are many intuitive ways to use $C_{vt}$ to guide correlation aggregation, such as addition, multiplication and concatenation, to enhance the semantic concept in the visual correlation map.
From our experiments in Tab.~\ref{tab:ablation_corr}, we find that simply updating $C_v^4$ by concatenating with $C'_{vt}$ together along the first dimension is efficient enough without introducing additional learnable parameters.
The combination of $C_v^4$ and $C'_{vt}$ is then accomplished multi-scale correlation aggregation.
The correlation aggregation architecture is commonly seen in recent FSS methods~\cite{min2021hypercorrelation,hong2022cost,xiong2022doubly}.
We introduce the center-pivot 4D convolution pyramid encoder (CCPE)~\cite{min2021hypercorrelation} to accomplish the correlation aggregation, which aggregates $\{C_v^i\}_{i=2}^4$ by employing a top-down approach through pyramidal processing.
In this process, the aggregation of lower-level correlation map $C_{v}^{i-1}$ is guided by higher-level correlation map $C_v^{i}$.
After the progressive update, $\{C_v^{i}\}_{i=2}^4$ is transformed to the final correlation map set $\{\hat{C}_v^{i}\}_{i=2}^4$, where $\hat{C}_v^{i} \in \mathbb{R}^{128 \times h_i \times w_i \times 2 \times 2}$ as in \cite{min2021hypercorrelation}.
And each of $\{\hat{C}_v^{i}\}_{i=2}^4$ is further compressed by average the last two dimensions to obtain the 2D feature map set $\hat{F}_q = \{\hat{f}_q^i\}_{i=2}^{4}$, where $\hat{f}_q^i \in \mathbb{R}^{128 \times h_i \times w_i}$.
Note the correlation aggregation method is only a plug-and-play component in our method.
As this operation is standard in FSS, it can be replaced with other methods, \eg, 4D swin transformer~\cite{hong2022cost} or deformable 4D transformer~\cite{xiong2022doubly}. For the sake of computational efficiency, we chose to use the CCPE in our experiments.

\subsection{Cross-modal Decoder}
\label{sec:decoder}

The cross-modal decoder is illustrated in  Fig.~\ref{fig:decoder} where the \eim is embedded to integrate the support and query embeddings.

\textbf{\Eim (\EIM).}
For fully exploiting the information between the aligned visual and textual embeddings, \ie, $f_v$ and $f_t$, we propose the EIM for visual-textual information mining.
The overall structure of the EIM is illustrated in the right of  Fig.~\ref{fig:decoder}.
The input $f_{v}$ undergoes an initial linear projection before being reshaped into the query vector $Q_v$.
Simultaneously, the input $f_{t}$ is converted into two distinct vectors: the key vector $K_t$ and the value vector $V_t$.
Then the multi-head attention (MHA)~\cite{vaswani2017attention} and linear layers are performed between these vectors to generate element-level attention weight $A_{vt}$.
$f_{v}$ is reduced dimension by the linear layer, multiplied with the attention weight, and forwarded through the cascaded linear layers to obtain enhanced feature map $f_{vt}$.
The process is formulated as follows:
\begin{equation}
  \begin{aligned}
    Q_v, K_t, V_t & = \mathrm{Linear}(f_{v}, f_{t}, f_{t}), \\
    A_{vt}        & = \mathrm{Linear}(\mathrm{MHA}(Q_v, K_t, V_t)), \\
    f_{vt}        & = \mathrm{Linear}(\mathrm{Linear}(f_{v}) \odot A_{vt}).
  \end{aligned}
\end{equation}

\textbf{Top-Down Decoding Path.}
Based on the feature map $f_{vt}$, the decoder gradually absorbs the 2D feature map set $\{\hat{f}_q^i\}_{i=2}^4$ from Sec. \ref{sec:ms_corragg} and the shallow feature maps of query after the linear transformation $\tilde{f}_q^1$ and $\tilde{f}_q^2$ through a top-down path.
In this process, low-resolution feature maps are progressively up-sampled by the bilinear interpolation, concatenated with high-resolution low-level feature maps, and then aggregated by stacked convolution blocks~\cite{wang2023internimage, he2016deep}.
Finally, a simple linear head is used to segment the region with the category of interest and generate the final prediction $P$.

\subsection{Supervision and $K$-Shot Inference.}
During the training phase, parameters are optimized by minimizing a combination of the cross-entropy loss and the dice loss~\cite{milletari2016v}.
In the context of $K$-shot inference, the network is provided with a set of support pairs $\mathcal{S}$ = $\{I_s^k, M_s^k, T_s^k\}_{k=1}^K$ along with a query image $I_q$.
It sequentially conducts $K$ individual one-shot inferences, yielding a collection of $K$ query predictions $\{P^k\}_{k=1}^K$. These predictions collaboratively contribute through a pixel-wise voting process to derive the final segmentation map, delineating the foreground and background.


\section{Experiments}
\label{sec:exper}

\begin{table*}[t]
  \centering
  \caption{Mask FSS and class-aware mask FSS performance comparison on PASCAL-5$^{i}$~\cite{shaban2017one} using mIoU (\%) and FB-IoU (\%) evaluation metrics. Numbers in \textbf{bold} indicate the best performance.}
  \resizebox{0.9\linewidth}{!}{
    \begin{tabular}{c|c|cccccc|cccccc}
  \toprule[1pt]
  \multirow{2}{*}{Backbone}   &
  \multirow{2}{*}{Methods}    &
  \multicolumn{6}{c|}{1-shot} &
  \multicolumn{6}{c}{5-shot}                                                                                                                                                                                                                                         \\
                              &                                      & $5^0$         & $5^1$         & $5^2$         & $5^3$         & mIoU          & FB-IoU        & $5^0$         & $5^1$         & $5^2$         & $5^3$         & mIoU          & FB-IoU        \\
  \midrule[1pt]
  \multicolumn{14}{c}{\textbf{Mask FSS}}  \\
  \midrule[1pt]
  \multirow{11}{*}{ResNet50}   & PFENet~\cite{tian2020prior}          & 61.7          & 69.5          & 55.4          & 56.3          & 60.8          & 73.3          & 63.1          & 70.7          & 55.8          & 57.9          & 61.9          & 73.9          \\
                              & HSNet~\cite{min2021hypercorrelation} & 64.3          & 70.7          & 60.3          & 60.5          & 64.0          & 76.7          & 70.3          & 73.2          & 67.4          & 67.1          & 69.5          & 80.6          \\
                              & IPRNet~\cite{okazawa2022interclass}  & 65.2          & 72.9          & 63.3          & 61.3          & 65.7          & -             & 70.2          & 75.6          & 68.9          & 66.2          & 70.2          & -             \\
                              & SSP~\cite{fan2022ssp}                & 60.5          & 67.8          & \textbf{66.4}          & 51.0          & 61.4          & -             & 67.5          & 72.3          & \textbf{75.2}          & 62.1          & 69.3          & -             \\
                              & DACM~\cite{xiong2022doubly}       & 66.5          & 72.6          & 62.2          & 61.3          & 65.7          & 77.8             & 72.4          & 73.7          & 69.1          & 68.4          & 70.9          & 81.3             \\
                              & DCAMA~\cite{shi2022dense}            & 67.5          & 72.3          & 59.6          & 59.0          & 64.6          & 75.7          & 70.5          & 73.9          & 63.7          & 65.8          & 68.5          & 79.5          \\
                              & VAT~\cite{hong2022cost}              & 67.6          & 72.0          & 62.3          & 60.1          & 65.5          & 77.8          & 72.4          & 73.6          & 68.6          & 65.7          & 70.1          & 80.9          \\
                              & FECANet~\cite{liu2023fecanet}     & 69.2          & 72.3          & 62.4          & 65.7          & 67.4          & 78.7          & 72.9          & 74.0          & 65.2          & 67.8          & 70.0          & 80.7          \\
                              & ABCNet~\cite{wang2023rethinking}     & 68.8          & 73.4          & 62.3          & 59.5          & 66.0          & 76.0          & 71.7          & 74.2          & 65.4          & 67.0          & 69.6          & 80.0          \\
                              & MSI~\cite{moon2023msi}               & 71.0          & 72.5          & 63.8          & 65.9          & 68.3          & 79.1          & 73.0          & 74.2          & 66.6          & \textbf{70.5}          & 71.1          & 81.2          \\
                              & Ours                                 & \textbf{72.7} & \textbf{75.6}          & 63.7          & \textbf{66.9} & \textbf{69.7}          & \textbf{80.1}          & \textbf{75.4} & \textbf{77.1}  & 67.9  & 69.9 & \textbf{72.6}  &  \textbf{82.3}         \\
  \midrule[1pt]
  \multicolumn{14}{c}{\textbf{Class-aware Mask FSS}}  \\
  \midrule[1pt]
  \multirow{2}{*}{ResNet50}   
                              & PGMANet~\cite{shuai2023pgmanet}      & 73.4          & 80.8          & 70.5          & 71.7          & 74.1          & 83.5          & 74.0          & 81.5          & 71.9          & 73.3          & 75.2          & 84.2          \\
                              & Ours                                 & \textbf{76.2} & 80.4          & 68.0          & \textbf{76.9} & 75.4          & 83.9          & \textbf{76.3} & 80.8          & 68.9          & \textbf{77.8} & 76.0          & 84.2          \\
    \midrule[1pt]
  \multirow{2}{*}{ViT-B/16}   
                              & PGMANet~\cite{shuai2023pgmanet}      & 74.0          & \textbf{81.9} & 66.8          & 73.7          & 74.1          & 82.1          & 74.5          & \textbf{82.2} & 67.2          & 74.4          & 74.6          & 82.5          \\
                              & Ours                                 & 75.0          & 79.6          & \textbf{74.7} & 76.4          & \textbf{76.4} & \textbf{85.3} & 75.5          & 79.9          & \textbf{75.9} & 77.5          & \textbf{77.2} & \textbf{86.0} \\
  \bottomrule[1pt]
\end{tabular}
  }
  \label{tab:pascal_sota}
\end{table*}

\begin{table*}[t]
  \centering
  \caption{Mask FSS and class-aware mask FSS performance comparison on COCO-20$^{i}$~\cite{lin2014microsoft} using mIoU (\%) and FB-IoU (\%) evaluation metrics. Numbers in \textbf{bold} indicate the best performance.}
  \resizebox{0.9\linewidth}{!}{
    \begin{tabular}{c|c|cccccc|cccccc}
  \toprule[1pt]
  \multirow{2}{*}{Backbone}   &
  \multirow{2}{*}{Methods}    &
  \multicolumn{6}{c|}{1-shot} &
  \multicolumn{6}{c}{5-shot}                                                                                                                                                                                                                                         \\
                              &                                      & $20^0$        & $20^1$        & $20^2$        & $20^3$        & mIoU          & FB-IoU        & $20^0$        & $20^1$        & $20^2$        & $20^3$        & mIoU          & FB-IoU        \\
  \midrule[1pt]
  \multicolumn{14}{c}{\textbf{Mask FSS}}  \\
  \midrule[1pt]
  \multirow{11}{*}{ResNet50}   & PFENet~\cite{tian2020prior}          & 36.5          & 38.6          & 34.5          & 33.8          & 35.8          & -             & 36.5          & 43.3          & 37.8          & 38.4          & 39.0          & -             \\
                              & HSNet~\cite{min2021hypercorrelation} & 36.3          & 43.1          & 38.7          & 38.7          & 39.2          & 68.2          & 43.3          & 51.3          & 48.2          & 45.0          & 46.9          & 70.7          \\
                              & IPRNet~\cite{okazawa2022interclass}  & 42.2          & 48.9          & 45.5          & 44.6          & 45.3          & -             & 48.0          & 55.7          & 50.7          & 50.1          & 51.1          & -             \\
                              & SSP~\cite{fan2022ssp}                & 35.5          & 39.6          & 37.9          & 36.7          & 37.4          & -             & 40.6          & 47.0          & 45.1          & 43.9          & 44.1          & -             \\
                              & DACM~\cite{xiong2022doubly}          & 37.5          & 44.3          & 40.6          & 40.1          & 40.6          & 68.9          & 44.6          & 52.0          & 49.2          & 46.4          & 48.1          & 71.6           \\
                              & DCAMA~\cite{shi2022dense}            & 41.9          & 45.1          & 44.4          & 41.7          & 43.3          & 69.5          & 45.9          & 50.5          & 50.7          & 46.0          & 48.3          & 71.7          \\
                              & VAT~\cite{hong2022cost}              & 39.0          & 43.7          & 42.6          & 39.7          & 41.3          & 68.8          & 44.1          & 51.1          & 50.2          & 46.1          & 47.9          & 72.4          \\
                              & FECANet~\cite{liu2023fecanet}        & 38.5          & 44.6          & 42.6          & 40.7          & 41.6          & 69.6          & 44.6          & 51.5          & 48.4          & 45.8          & 47.6          & 71.1          \\
                              & ABCNet~\cite{wang2023rethinking}     & 42.3          & 46.2          & 46.0          & 42.0          & 44.1          & 69.9          & 45.5          & 51.7          & 52.6          & 46.4          & 49.1          & 72.7          \\
                              & MSI~\cite{wang2023rethinking}        & 42.4          & 49.2          & \textbf{49.4}          & 46.1          & 46.8          & -             & 47.1          & 54.9          & 54.1          & 51.9          & 52.0          & -             \\
                              & Ours                                 & \textbf{46.5} & \textbf{53.0}          & 48.0          & \textbf{48.2} & \textbf{48.9} & \textbf{72.4}          & \textbf{50.3} & \textbf{59.5} & \textbf{54.4} & \textbf{52.0} & \textbf{54.1} & \textbf{75.0} \\
  \midrule[1pt]
  \multicolumn{14}{c}{\textbf{Class-aware Mask FSS}}  \\
  \midrule[1pt]
  \multirow{2}{*}{ResNet50}
                              & PGMANet~\cite{shuai2023pgmanet}      & 49.9          & 56.7          & 55.8          & 54.7          & 54.3          & 75.8          & 49.5          & 61.7          & 59.1          & 57.9          & 57.1          & 76.7          \\
                              & Ours                                 & \textbf{52.6} & 59.8          & 57.6          & \textbf{56.8} & \textbf{56.7} & 76.8          & 52.3          & \textbf{62.4}          & \textbf{60.8} & \textbf{57.0} & \textbf{58.1} & \textbf{77.8} \\
    \midrule[1pt]
  \multirow{1}{*}{ViT-B/16}   
                              & Ours                                 & 51.2          & \textbf{61.8} & \textbf{58.0} & 55.6          & \textbf{56.7} & \textbf{77.0} & \textbf{53.1} & \textbf{62.4} & 59.2          & 56.8          & 57.9          & \textbf{77.8} \\
  \bottomrule[1pt]
\end{tabular}

  }
  \label{tab:coco_sota}
\end{table*}

\subsection{Experimental Settings}
\label{sec:exp}

\textbf{Datasets.}
We conduct experiments on three commonly used FSS benchmark datasets, \ie, PASCAL-5$^i$~\cite{shaban2017one}, COCO-20$^i$~\cite{lin2014microsoft}, and FSS-1000~\cite{li2020fss1000} to evaluate our method.
PASCAL-5$^i$ contains PASCAL VOC 2012~\cite{everingham2010pascal} with additional mask annotations~\cite{hariharan2014simultaneous}.
It contains 20 classes, which are evenly split into four folds for cross-validation, \ie, three of which for training and the rest for testing.
COCO-20$^i$ is generated from MS-COCO~\cite{lin2014microsoft} and contains 80 categories.
The dataset is evenly divided into 4 folds of 20 classes each fold as done for PASCAL-5$^i$.
FSS-1000 contains 1000 object categories, each of which includes multiple binary-annotated images. Following~\cite{li2020fss1000}, the 1000 categories in the FSS-1000 are divided into 3 splits for training, validation, and testing, with each split containing 520, 240, and 240 categories, respectively.
All images in the above three datasets are natural images. To validate the generalization capability of our method, we conducted experiments on iSAID-5$^i$~\cite{yao2021sdm}, a few-shot remote sensing segmentation dataset. iSAID-5$^i$ is modified from iSAID~\cite{waqas2019isaid}, which contains 15 categories across 2,806 high-resolution images. The 15 categories are divided into three folds for cross-validation.

\textbf{Evaluation metrics.}
Following ~\cite{min2021hypercorrelation,peng2023hierarchical,moon2023msi}, we adopt mean intersection over union
(mIoU) and foreground-background IoU (FB-IoU) as our evaluation metrics.
The main evaluation metric mIoU averages the IoU values for all classes in each fold.
The supplemental evaluation metric FB-IoU calculates the foreground and background IoU values ignoring object classes and averages them.

\subsection{Implementation Details}
\label{sec:imp_details}

All experiments are implemented in PyTorch~\cite{paszke2019pytorch} and optimized using AdamW~\cite{AdamW} with a fixed learning rate of 0.0005.
The spatial resolutions of both support and query images are set to $400 \times 400$ throughout all experiments.
CATs data augmentation~\cite{cho2021cats,hong2022cost} is used and the batch size for all experiments is set to 16.
For the backbone network, we use pre-trained ResNet50~\cite{he2016deep} and ViT-B/16~\cite{dosovitskiy2020image} from the contrastive language-image model CLIP~\cite{radford2021clip} like the existing methods~\cite{shuai2023pgmanet,lueddecke22clipseg}. 
In concordance with the protocol of recent studies~\cite{min2021hypercorrelation,hong2022cost,peng2023hierarchical,moon2023msi}, we keep the weights of the backbone network frozen during the training process.
Additionally, we bypass complex prompt engineering by directly inputting support class text into the CLIP textual encoder. 
We conduct all experiments on a computer with an i7-10700F CPU and a single RTX 3090 GPU. It takes about 17 hours per fold training on PASCAL-5$^i$~\cite{shaban2017one} and 40 hours per fold training on COCO-20$^i$~\cite{lin2014microsoft}.

\begin{table*}[t]
  \centering
  \caption{Task patterns performance comparison on PASCAL-5$^{i}$~\cite{shaban2017one} using mIoU (\%) and FB-IoU (\%) evaluation metrics.
  Numbers in \textbf{bold} indicate the best performance.
  }
  \resizebox{0.9\linewidth}{!}{

\begin{tabular}{c|c|cccccc|cccccc}
  \toprule[1pt]
                                        &     & \multicolumn{6}{c|}{1-shot} & \multicolumn{6}{c}{5-shot}                                                                                                                                                                 \\
  \multirow{-2}{*}{Methods} & \multirow{-2}{*}{Backbone}  & $5^0$                       & $5^1$                      & $5^2$         & $5^3$         & mIoU          & FB-IoU        & $5^0$         & $5^1$         & $5^2$         & $5^3$         & mIoU          & FB-IoU        \\
  \midrule[1pt]
  \multicolumn{14}{c}{\textbf{Box FSS}}   \\
  \midrule[1pt]
  HSNet~\cite{min2021hypercorrelation} & ResNet50 & 56.8                        & 67.0                       & 54.8          & 55.7          & 58.6          & 72.3          & 65.1          & 71.1          & 62.8          & 63.8          & 65.7          & 77.4          \\
  DCAMA~\cite{shi2022dense} & ResNet50  & 64.2                        & 70.7                       & 57.4          & 57.2          & 62.4          & 74.0          & 70.5          & 73.9          & 63.7          & 65.8          & 68.5          & 79.5          \\
  Ours & ResNet50 & \textbf{72.3} & \textbf{73.9} & \textbf{62.4} & \textbf{66.2} & \textbf{68.7} & \textbf{78.9} & \textbf{74.8} & \textbf{76.1} & \textbf{66.8} & \textbf{69.3} & \textbf{71.8} & \textbf{81.3}   \\
  \midrule[1pt]
  \multicolumn{14}{c}{\textbf{Class-aware Box FSS}}   \\
  \midrule[1pt]
  PGMANet~\cite{shuai2023pgmanet} & ResNet50  & 72.4                        & \textbf{80.8}              & \textbf{68.9} & 70.7          & 73.2          & 82.5          & 73.1          & \textbf{81.5} & \textbf{69.8} & 72.1          & 74.1          & 83.1          \\
  Ours  & ResNet50 & \textbf{75.6}               & 80.4                       & 67.7          & \textbf{75.9} & \textbf{74.7} & \textbf{83.3} & \textbf{76.0} & 80.8          & 69.5          & \textbf{77.0} & \textbf{75.6} & \textbf{83.7} \\
  \midrule[1pt]
  \multicolumn{14}{c}{\textbf{Image FSS}}  \\
  \midrule[1pt]
  HSNet~\cite{min2021hypercorrelation}& ResNet101 & \textbf{66.2}                      & {69.5}                     & {53.9}        & \textbf{56.2}        & {61.5}        & 72.5          & \textbf{68.9}        & \textbf{71.9}          & 56.3          & \textbf{57.9}          & 63.7          & 73.8          \\
  Ours  & ResNet50  & 65.7 & \textbf{70.2} & \textbf{55.0} & \textbf{56.2} & \textbf{61.8} & \textbf{73.3} & \textbf{68.9} & 71.6 & \textbf{57.6} & 57.5 & \textbf{63.9} & \textbf{74.4} \\
  \midrule[1pt]
  \multicolumn{14}{c}{\textbf{Class-aware Image FSS}}  \\
  \midrule[1pt]
  (V+S)-1~\cite{siam2020weakly} & ResNet50 & 49.5                        & 65.5                       & 50.0          & 49.2          & 53.5          & 65.6          & -             & -             & -             & -             & -             & -             \\
  (V+S)-2~\cite{siam2020weakly} & ResNet50 & 42.5                        & 64.8                       & 48.1          & 46.5          & 50.5          & 64.1          & 45.9          & 65.7          & 48.6          & 46.6          & 51.7          & -             \\
  IMR-HSNet~\cite{wang2022imr} & ResNet50 & {62.6}                      & {69.1}                     & {56.1}        & {56.7}        & {61.1}        & -             & -             & -            & -             & -             & -             & -             \\
  WRCAM~\cite{zhang2022weakly} & ResNet101 & {58.1}                      & {65.0}                     & {60.3}        & {49.0}        & {58.1}        & -             & -             & -             & -             & -             & -             & -             \\
  PGMANet~\cite{shuai2023pgmanet}& ResNet50& 68.6                        & 76.3                       & 60.3          & 64.1          & 67.3          & 77.8          & 68.9          & 76.6          & 60.5          & 64.1          & 67.5          & 78.0          \\
  Ours & ResNet50 & \textbf{71.3}               & \textbf{77.0}              & \textbf{62.6} & \textbf{69.4} & \textbf{70.1} & \textbf{79.8} & \textbf{73.1} & \textbf{78.6} & \textbf{66.7} & \textbf{71.0} & \textbf{72.3} & \textbf{80.6} \\
  \midrule[1pt]
  \multicolumn{14}{c}{\textbf{Text FSS}}                                                                                                                                                                                                                \\
  \midrule[1pt]
  SPNet~\cite{xian2019semantic} & ResNet101 & 23.8                        & 17.0                       & 14.1          & 18.3          & 18.3          & 44.3          & -             & -             & -             & -             & -             & -             \\
  ZS3Net~\cite{bucher2019zero} & ResNet101 & 40.8                        & 39.4                       & 39.3          & 33.6          & 38.3          & 57.7          & -             & -             & -             & -             & -             & -             \\
  LSeg~\cite{li2022languagedriven}& ResNet101 & 52.8                        & 53.8                       & 44.4          & 38.5          & 47.4          & 64.1          & -             & -             & -             & -             & -             & -             \\
  LSeg~\cite{li2022languagedriven}& ViT-L/16 & 61.3                        & 63.6                       & 43.1          & 41.0          & 52.3          & 67.0          & -             & -             & -             & -             & -             & -             \\
  SAZS~\cite{liu2023delving} & ResNet101 & 57.3                        & 60.3                       & 58.4          & 45.9          & 55.5          & 66.4          & -             & -             & -             & -             & -             & -             \\
  SAZS~\cite{liu2023delving} & ViT-L/16 & 62.7                        & 64.3                       & \textbf{60.6}          & 50.2          & 59.4          & 69.0          & -             & -             & -             & -             & -             & -             \\
  PGMANet~\cite{shuai2023pgmanet} & ResNet50 & 54.2                        & 67.7                       & 57.5          & 60.9          & 60.1          & 72.1          & -             & -             & -             & -             & -             & -             \\
  Ours & ResNet50 & \textbf{67.3}               & \textbf{71.8}              & 59.3 & \textbf{66.0} & \textbf{66.1} & \textbf{76.2} & -             & -             & -             & -             & -             & -             \\
  \bottomrule[1pt]
\end{tabular}

  }
  \label{tab:add_tasks}
\end{table*}

\subsection{Task Patterns}
\label{sec:task_patterns}

\begin{figure*}[t]
  \centering
  \includegraphics[width=\linewidth]{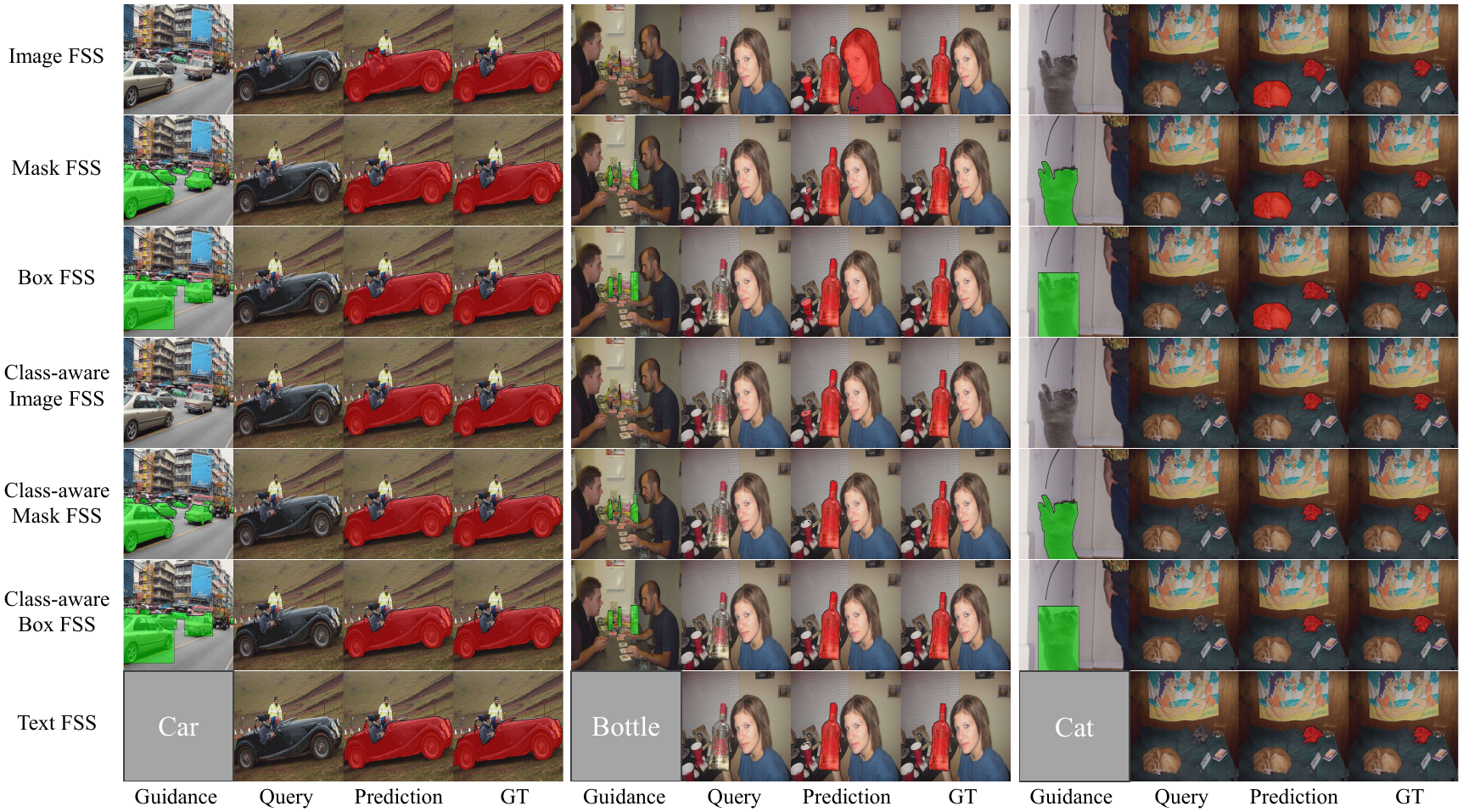}
  \caption{7 task patterns with different guidance types. For simplicity, we omit the text information in the guidance types of class-aware FSS.}
  \label{fig:pattern}
\end{figure*}

As summarised in~ Fig.~\ref{fig:Figure1}, the FSS field be categorized into seven distinct patterns based on the form of guidance.
Accordingly, we introduce the UniFSS framework for a universal examination of these diverse patterns.
Previous sections detail the framework with the example of class-aware image FSS.
In the box patterns, \ie, \ding{174} and \ding{177}, a binary mask can be created by encapsulating a rectangular region within the box, which serves as $M_s$.
For the pattern without the class text input, \ie \ding{172}-\ding{174}, $f_t$ can be embedded from the deep feature map of support image $I_s$ after being masked average pooled by the binary mask $M_s$ in mask and box FSS or global average pooled in image FSS.
Especially, in text FSS, \ie, \ding{178}, the support set is exclusively composed of class text.
For acquiring support visual features, the query image $I_q$ is treated as the support image $I_s$ itself. In Fig.~\ref{fig:pattern}, we illustrated the guidance types in different patterns.

We find that training model parameters only on a single task pattern doesn't achieve optimal results across all task patterns. Sampling different guidance types for training the model can alleviate this problem. However, we believe that sampling different guidance types during training is too cumbersome and impractical for real-world applications. Therefore, we group the task patterns and train a set of parameters for each group of patterns. Note that the model architecture remains unchanged for all task patterns.
In our experiments, we train three sets of model parameters to accomplish seven task patterns, \ie, 1) training on \ding{172} and testing on \ding{172}, 2) training on \ding{173} and testing on \ding{173} and \ding{174}, 3) training on \ding{176} and testing on \ding{175}-\ding{178} following~\cite{shuai2023pgmanet}.

\subsection{Comparison with State-of-the-Art Methods}

\textbf{Mask FSS and Class-aware Mask FSS.}
Both settings are more common, so we compare our UniFSS with recent state-of-the-art approaches on PASCAL-5$^i$~\cite{shaban2017one} and COCO-20$^i$~\cite{lin2014microsoft} in Tab.~\ref{tab:pascal_sota} and Tab.~\ref{tab:coco_sota}.
It can be observed that the proposed method demonstrates a significant advancement over previous advanced approaches, achieving new state-of-the-art results in terms of mIoU and FB-IoU.
For PASCAL-5$^i$, our approach with ResNet50 achieves relative mIoU improvement of 2.0\% and 2.1\% under the $1$-shot and $5$-shot setting in mask FSS, respectively.
As for class-aware mask FSS, our ResNet50 and ViT-B/16 versions surpass the recent best practice PGMANet~\cite{shuai2023pgmanet} by 1.8\% and 3.1\% in terms of relative mIoU, respectively.
For COCO-20$^i$, our method with ResNet50 outperforms the state-of-the-art methods by 4.5\% and 4.4\% in terms of relative mIoU, respectively.

We do not include the methods MIANet~\cite{yang2023mianet} and HDMNet~\cite{peng2023hierarchical} in Tab.~\ref{tab:pascal_sota} and Tab.~\ref{tab:coco_sota}. The reasons are twofold: \lowercase\expandafter{\romannumeral1}) They employ a two-stage training strategy, where they first train a semantic segmentation model on base categories, and then freeze this model to train a FSS model. During testing, they utilize the prediction maps of base categories to suppress incorrect results for novel categories. This setting is unfair when compared to the previous FSS methods. 

\textbf{Box FSS and Class-aware Box FSS.}
From Tab.~\ref{tab:add_tasks}, it can be observed that our method outperforms these competitors by 10.1\% and 2.0\%  under different patterns in terms of relative mIoU, respectively, indicating its robustness under weak annotation.

\textbf{Image FSS and Class-aware Image FSS.}
Such a task pattern aims at segmenting targets with the common category in several images, which is a more challenging setting.
The results are summarized in Tab.~\ref{tab:add_tasks}.
For class-aware image FSS, our method achieves the relative mIoU improvement of 14.7\% compared to IMR-HSNet~\cite{wang2022imr}, which is designed specifically for this setting.
When the support mask is removed, our method exhibits a smaller performance drop compared to PGMANet (7.0\% \versus 9.2\%).
As for image FSS, our method obtains comparable results against previous works~\cite{min2021hypercorrelation}.

\textbf{Text FSS.}
In this task pattern, only category information is utilized as the support information to segment query images.
Our method can be directly extended to this setting by replacing support images with query image as mentioned in~Sec. \ref{sec:task_patterns}.
The results in Tab.~\ref{tab:add_tasks} demonstrate our approach outperforms all state-of-the-art methods.
As for the performance drop after removal of the support mask, our method exhibits a lower decrease compared to PGMANet (12.3\% \versus 18.8\%).

\begin{table}[t]
  \centering
  \caption{1-shot and 5-shot performance comparison with FSS and SAM-series methods on FSS-1000~\cite{li2020fss1000} under the mask FSS setting using mIoU(\%) evaluation metric. * indicates under the class-aware mask FSS setting.}
  \resizebox{\linewidth}{!}{
    \begin{tabular}{c|c|cc}
  \toprule[1pt]
  Methods    & Backbone & 1-shot & 5-shot          \\
  \midrule[1pt]
  FSOT~\cite{liu2022fsot}   &  ResNet50 & 82.5 & 83.8         \\
  HSNet~\cite{min2021hypercorrelation}   &  ResNet50 & 85.5 & 87.8         \\
  MSI+HSNet~\cite{moon2023msi}   &  ResNet50 & 87.5 & 88.4          \\
  PerSAM~\cite{zhang2023personalize}   &  ViT-H & 71.2 & -         \\
  Matcher~\cite{liu2023matcher}  &  ViT-H \& DINOv2  & 87.0 & -         \\
  Ours     &  ResNet50 & 89.8 & 90.2 \\
  Ours*     &  ResNet50 & 89.6 & 90.0 \\
  \bottomrule[1pt]
\end{tabular}

  }
  \label{tab:fss1000_sota}
\end{table}

\begin{table}[t]
  \centering
  \caption{1-shot performance comparison with other methods on iSAID-5$^i$~\cite{yao2021sdm} under the mask FSS setting using mIoU(\%) evaluation metric.}
  \resizebox{\linewidth}{!}{
    \begin{tabular}{c|c|cccc}
  \toprule[1pt]
  Methods    & Backbone & $5^0$         & $5^1$         & $5^2$        & mIoU          \\
  \midrule[1pt]
  SDM~\cite{yao2021sdm}   &  ResNet50 & 34.3          & 22.3          & 35.6            & 30.7          \\
  HSNet~\cite{min2021hypercorrelation}   &  ResNet50 & 30.8          & 24.4          & 38.2            & 31.1          \\
  SCCNet~\cite{liu2023matcher}  &  ResNet50  & 36.2          & 27.4          & 43.4          &  35.7          \\
  Ours     &  ResNet50 & 36.9 & 29.4 & 37.2  & 34.5 \\
  \bottomrule[1pt]
\end{tabular}
  }
  \label{tab:isaid_sota}
\end{table}

\begin{table}[t]
  \centering
  \caption{Comparison with SAM-series methods on COCO-20$^i$~\cite{lin2014microsoft}. * indicates under the class-aware mask FSS setting.}
  \resizebox{\linewidth}{!}{
    \begin{tabular}{c|c|ccccc}
  \toprule[1pt]
  Methods    & Backbone & $20^0$         & $20^1$         & $20^2$         & $20^3$         & mIoU          \\
  \midrule[1pt]
  PerSAM~\cite{zhang2023personalize}   &  ViT-H & 23.1          & 23.6          & 22.0          & 23.4         & 23.0          \\
  Matcher~\cite{liu2023matcher}  &  ViT-H \& DINOv2  & \textbf{52.7}          & 53.5          & 52.6          & 52.1          & 52.7          \\
  Ours     &  ResNet50 & 46.5 & 53.0 & 48.0 & 48.2 & 48.9 \\
  Ours*   &  ViT-B & 51.2 & \textbf{61.8} & \textbf{58.0} & \textbf{55.6} & \textbf{56.7} \\
  \bottomrule[1pt]
\end{tabular}

  }
  \label{tab:sam}
\end{table}

\begin{figure}[t]
  \centering
  \includegraphics[width=\linewidth]{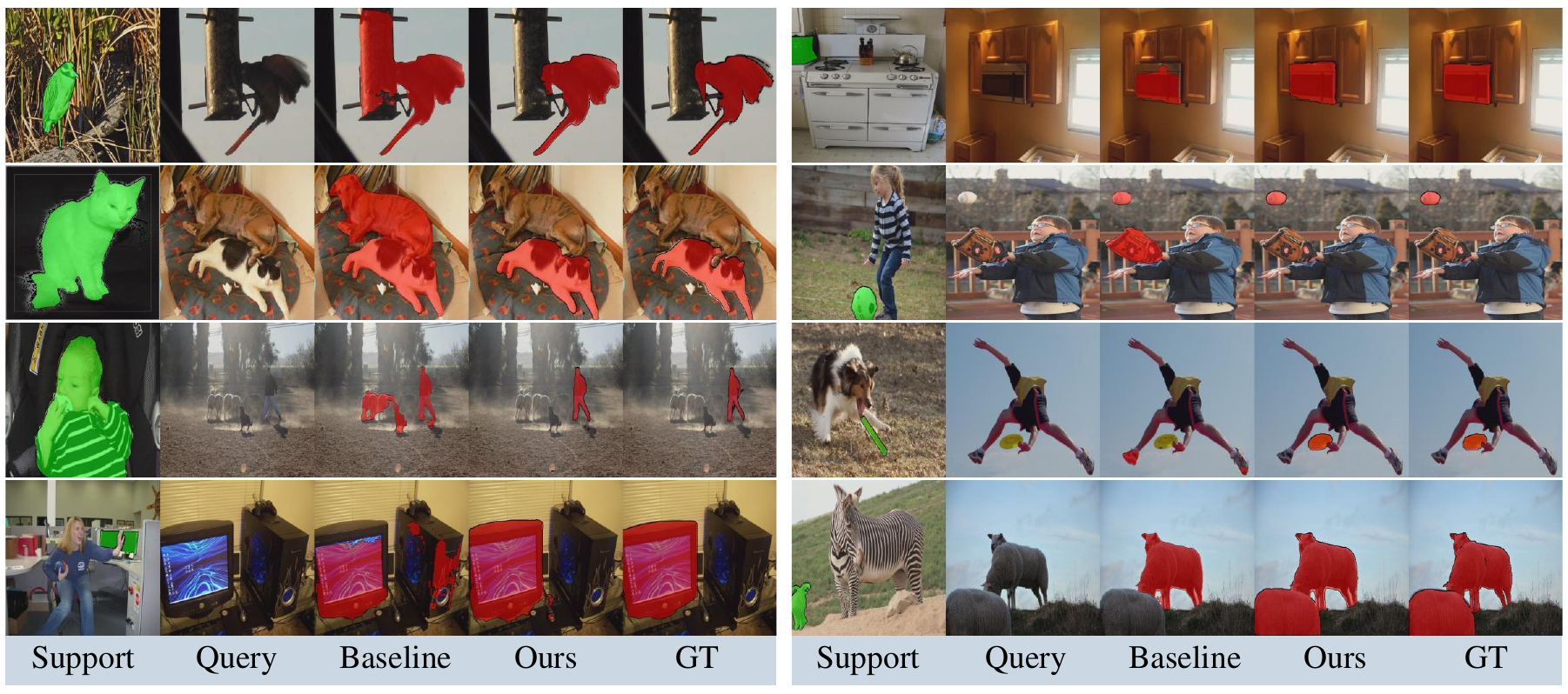}
  \caption{Qualitative comparison with the baseline on some samples from PASCAL-$5^i$ (\textit{Left}) and COCO-$20^i$ (\textit{Right}).
  UniFSS can better overcome the intra-class diversities.}
  \label{fig:prediction}
\end{figure}

\begin{figure}[t]
  \centering
  \includegraphics[width=\linewidth]{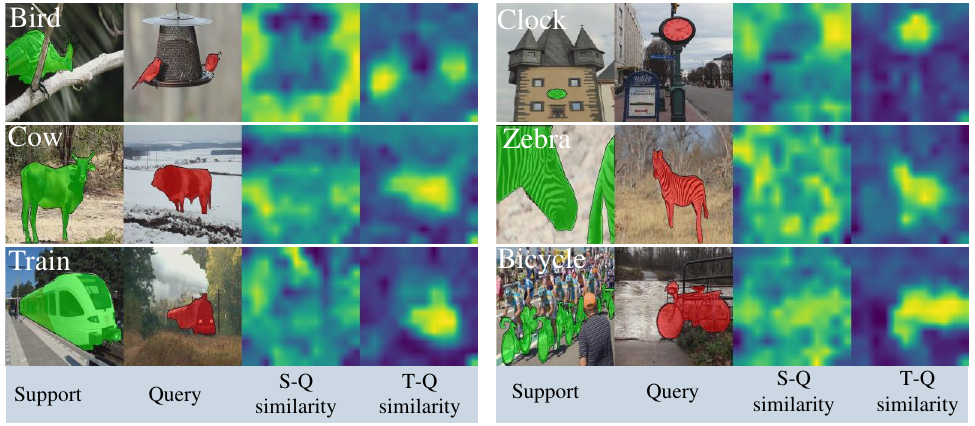}
  \caption{Visualization of support-query (S-Q) and text-query (T-Q) cosine similarity using our encoder on random samples from PASCAL-$5^i$ (\textit{Left}) and COCO-$20^i$ (\textit{Right}).
  T-Q similarity can provide accurate target localization.}
  \label{fig:visual_tq}
\end{figure}

\begin{figure}[t]
  \centering
  \includegraphics[width=\linewidth]{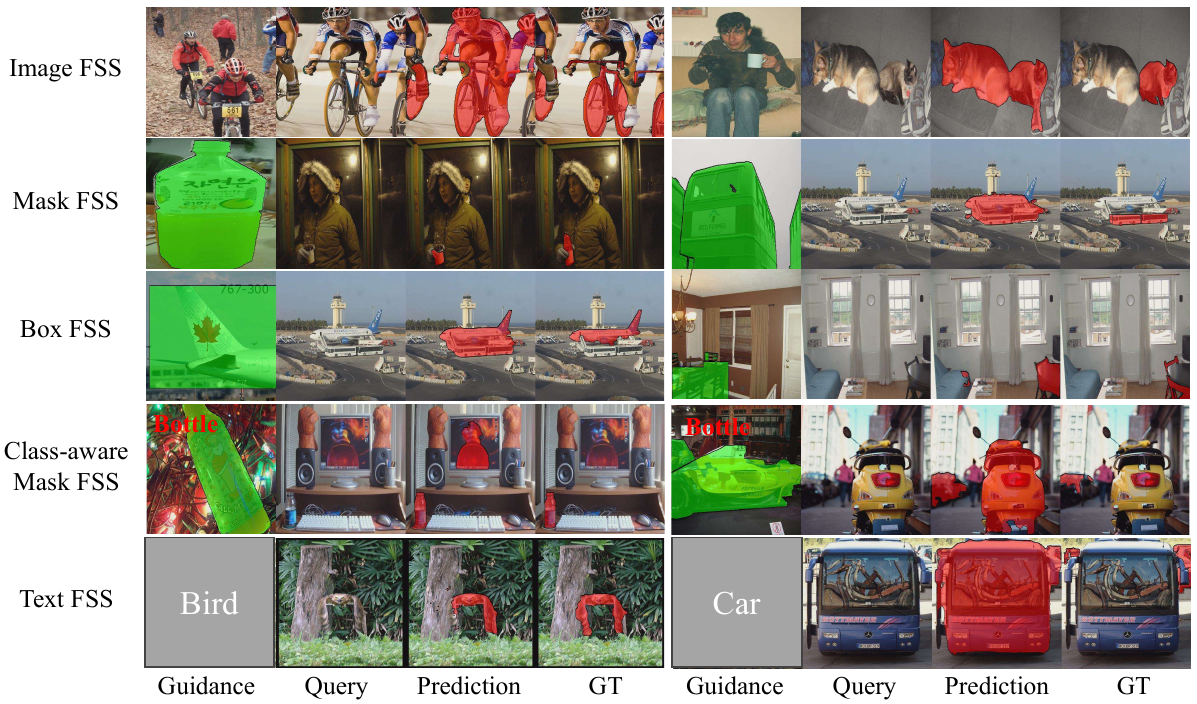}
  \caption{Visualization of some typical failure cases of different task patterns.}
  \label{fig:fail_case}
\end{figure}

\begin{table*}[!t]
  \centering
  \begin{minipage}[t]{0.57\linewidth} 
    \centering
    \caption{Ablation study for each component under class-aware mask FSS.
    \CorrOps: Sec. \ref{sec:corr_computation}.
  \HSCU: Sec. \ref{sec:hscu}.
  \EIM: Fig.~\ref{fig:decoder}.
  Best results are shown in \textbf{bold}.}
    \resizebox{1\linewidth}{!}{ 
      \begin{tabular}{ccc|ccccc|ccccc}
  \toprule[1pt]
  \multicolumn{3}{c|}{Components}
             &
  \multicolumn{5}{c|}{PASCAL-5$^i$}
             &
  \multicolumn{5}{c}{COCO-20$^i$}                                                                                                                                                                      \\
  \CorrOps        & \HSCU       & \EIM        & $5^0$         & $5^1$         & $5^2$         & $5^3$         & mIoU          & $20^0$        & $20^1$        & $20^2$        & $20^3$        & mIoU          \\
  \midrule[1pt]
             &            &            & 70.4          & 73.5          & 65.8          & 66.2          & 69.0          & 46.5          & 49.6          & 48.7          & 49.4          & 48.6          \\
  \checkmark &            &            & 73.5          & 78.0          & 63.8          & 73.0          & 72.1          & 48.0          & 57.1          & 54.2          & 53.6          & 53.2          \\
  \checkmark & \checkmark &            & 75.5          & 79.5          & 64.7          & 74.4          & 73.5          & 51.1          & 58.4          & 55.3          & 55.3          & 55.0          \\
  \checkmark &            & \checkmark & 74.8          & 79.9          & 65.9          & 74.7          & 73.8          & 50.3          & 58.3          & 55.4          & 54.2          & 54.6          \\
  \checkmark & \checkmark & \checkmark & \textbf{76.2} & \textbf{80.4} & \textbf{68.0} & \textbf{76.9} & \textbf{75.4} & \textbf{52.6} & \textbf{59.8} & \textbf{57.6} & \textbf{56.8} & \textbf{56.7} \\
  \bottomrule[1pt]
\end{tabular}

    }
    \label{tab:ablation_module}
  \end{minipage}
  \hfill
  \begin{minipage}[t]{0.4\linewidth} 
    \centering
    \caption{Ablation study on methods of building correlation map on PASCAL-5$^i$~\cite{shaban2017one} under class-aware mask FSS. Best results are shown in \textbf{bold}.}
    \resizebox{1\linewidth}{!}{ 
      \begin{tabular}{c|ccccc}
  \toprule[1pt]
  Methods              & $5^0$         & $5^1$         & $5^2$         & $5^3$         & mIoU          \\
  \midrule[1pt]
  Addition                  & 75.0          & 78.2          & 64.5          & 72.3          & 72.5          \\
  Multiplication             & 75.5          & 79.1          & 63.1          & 74.7          & 73.1          \\
  Concatenation (Ours) & \textbf{76.2} & \textbf{80.4} & \textbf{68.0} & \textbf{76.9} & \textbf{75.4} \\
  \bottomrule[1pt]
\end{tabular}

    }
    \label{tab:ablation_corr}
  \end{minipage}
\end{table*}

\begin{table*}[t]
  \centering
  \caption{Comparison of different guidance types from the support set.}
  \resizebox{\linewidth}{!}{
    \begin{tabular}{l|cccc|ccccc|ccccc}
  \toprule[1pt]
                                   &
  \multicolumn{4}{c|}{Guidance}
                                   &
  \multicolumn{5}{c|}{PASCAL-5$^i$}
                                   &
  \multicolumn{5}{c}{COCO-20$^i$}                                                                                                                                         \\
  Type                             & Image $I_I$ & Bbox $I_B$ & Mask $I_M$ & Text $I_T$ & $5^0$ & $5^1$ & $5^2$ & $5^3$ & mIoU & $20^0$ & $20^1$ & $20^2$ & $20^3$ & mIoU \\
  \midrule
  \ding{172} Image FSS             & \checkmark  &            &            &            & 65.7  & 70.2  & 55.0  & 56.2  & 61.8 & 30.8  & 33.9 & 33.6 & 30.7 & 32.3  \\
  \ding{173} Mask FSS              & \checkmark  &            & \checkmark &            & 72.7  & 75.6  & 63.7  & 66.9  & 69.7 & 46.5   & 53.0   & 48.0   & 48.2   & 48.9 \\
  \ding{174} BBox FSS              & \checkmark  & \checkmark &            &            & 71.8  & 73.8  & 62.4  & 66.3  & 68.6 & 43.7   & 49.6   & 41.5   & 44.4   & 44.8 \\
  \ding{175} Class-aware Image FSS & \checkmark  &            &            & \checkmark & 71.3  & 77.0  & 62.6  & 69.4  & 70.1 & 48.5   & 49.9   & 44.3   & 46.6   & 47.3 \\
  \ding{176} Class-aware Mask FSS  & \checkmark  &            & \checkmark & \checkmark & 76.2  & 80.4  & 68.0  & 76.9  & 75.4 & 52.6   & 59.8   & 57.6   & 56.8   & 56.7 \\
  \ding{177} Class-aware Box FSS   & \checkmark  & \checkmark &            & \checkmark & 75.6  & 80.4  & 67.7  & 75.9  & 74.7 & 51.6   & 57.9   & 55.5   & 55.8   & 55.2 \\
  \ding{178} Text FSS              &             &            &            & \checkmark & 67.3  & 71.8  & 59.3  & 66.0  & 66.1 & 44.1   & 42.4   & 40.6   & 40.0   & 41.8 \\
  \bottomrule[1pt]
\end{tabular}

  }
  \label{tab:input_compare}
\end{table*}

\textbf{Performance comparison on FSS-1000~\cite{li2020fss1000}.}  
In Tab.~\ref{tab:fss1000_sota}, we present the comparison of our method with other methods on FSS-1000~\cite{li2020fss1000} under the mask FSS and class-aware mask FSS setting. It can be observed that our method outperforms the other two methods~\cite{min2021hypercorrelation,moon2023msi} which also use center-pivot 4D convolution pyramid decoder~\cite{min2021hypercorrelation} under the mask FSS setting.
Note that on FSS-1000~\cite{li2020fss1000}, class-aware mask FSS has lower performance compared to mask FSS. This contradicts the conclusions drawn from PASCAL-5$^i$~\cite{shaban2017one} and COCO-20$^i$~\cite{lin2014microsoft}, where introducing text as guidance can significantly improve performance. This is because FSS-1000 contains many uncommon categories. Text embeddings cannot provide accurate target localization for these categories.

\textbf{Performance comparison on iSAID-5$^i$~\cite{yao2021sdm}.} In Tab.~\ref{tab:isaid_sota}, we present the 1-shot performance comparison on iSAID-5$^i$~\cite{yao2021sdm} under the mask FSS setting. 
Our method outperforms SDM~\cite{yao2021sdm} and achieves comparable results with SCCNet~\cite{wang2023sccnet}. Notably, both SDM~\cite{yao2021sdm} and SCCNet~\cite{wang2023sccnet} are specifically designed for few-shot remote sensing segmentation. This shows the generalizability of our method.

\textbf{Comparison with SAM-series methods.} 
In the era of SAM~\cite{kirillov2023segment}, some methods~\cite{zhang2023personalize,liu2023matcher} employ the combination of SAM with matching strategy or auxiliary models~\cite{oquab2023dinov2} to achieve one-shot segmentation.
Tab.~\ref{tab:sam} and Tab.~\ref{tab:fss1000_sota} show the comparison with SAM-series methods~\cite{zhang2023personalize,liu2023matcher}. It can be observed our method get comparable results with Matcher~\cite{liu2023matcher}.
Notably, Matcer~\cite{liu2023matcher} employs both ViT-H and DINOv2~\cite{oquab2023dinov2} as its backbone, which results in a substantial number of parameters and computational demands.
SAM-series models are currently not equipped to handle settings when text and image serve as guidance.
In comparison to our method, the advantages of SAM-series are that some of them are training-free~\cite{liu2023matcher} and are capable of segmenting with prompts from query images in the form of points, lines, and other structures with the help of SAM~\cite{kirillov2023segment}.

\subsection{Ablation Study}
\label{sec:ablation}

In this section, we first detail the contribution of each component and design to the overall network. Then, we compare the impact of different guidance types of support set on performance.
If not specified, all ablation experiments are conducted under $1$-shot setting of class-aware mask FSS with ResNet50 as the backbone.

\textbf{Ablation on backbone.}
Tab.~\ref{tab:pascal_sota} and Tab.~\ref{tab:coco_sota} demonstrate that our method achieves state-of-the-art results on two  common task patterns, mask FSS and class-aware mask FSS, whether using CNN-base or transformer-base backbone, which shows the robustness of our approach against different backbones.

\textbf{Effectiveness of components.}
We quantitatively show the benefit of each component in Tab.~\ref{tab:ablation_module}.
The model variant without
\hscu (\HSCU, Sec. \ref{sec:hscu}),
\corrops (\CorrOps, Sec. \ref{sec:corr_computation}),
and \eim (\EIM, Sec. \ref{sec:decoder})
is considered as the baseline.
Specifically, removing or adding \CorrOps means whether the visual and textual correlation is contained in our framework.
We add \CorrOps to the baseline model and the relative mIoU performance is significantly improved with the gain of 4.5\% and 9.5\% on PASCAL-5$^i$ and COCO-20$^i$, respectively.
%
The model performance is consistently improved after introducing the \HSCU and \EIM on top of this model.
When the \CorrOps, \HSCU, and \EIM are all added to the baseline model, the relative mIoU performance gains 9.3\% and 16.7\% on PASCAL-5$^i$ and COCO-20$^i$, respectively.
%
Fig.~\ref{fig:prediction} illustrates the qualitative comparison of UniFSS with the baseline network. When targets of the same class in query and support images have large intra-class diversities, the baseline network produces various degrees of false predictions. In contrast, UniFSS overcomes these problems and accurately predicts the targets.

\textbf{How to build correlation maps.}
 Fig.~\ref{fig:visual_tq} illustrates some support-query (S-Q) and text-query (T-Q) cosine similarity maps from our feature encoder.
It can be seen that T-Q similarity can provide accurate target localization.
So in Sec. \ref{sec:corr_computation}, we broadcast the initial visual-textual (\ie, T-Q) correlation map $C_{vt}$ to the 4D correlation and directly inject it into the deepest $C_v^{4}$ to introduce important object clues into the visual correlation map.
To explore better ways of injecting information, we experiment with different ways to build the deepest correlation map.
In Tab.~\ref{tab:ablation_corr}, we explored each of the three operations, including addition, multiplication, and the concatenation we used.
For the correlation map $C_{vt}$ and $C_v^{4}$, the addition operation means the element-wise addition, 
the multiplication operation refers to fusing the two correlation maps with the Hadamard product.
Our approach, \ie, concatenating them, shows the best results. Our method introduces only a few additional parameters. In mask FSS, box FSS, and image FSS, the text embeddings are replaced with the mask average pooling of the support features.

\textbf{The selection of multi-scale aggregation unit.} In Sec. \ref{sec:ms_corragg}, we take center-pivot 4D convolution pyramid decoder (CCPE)~\cite{min2021hypercorrelation} as multi-scale aggregation unit because of its computational efficiency. With CCPE, we can set the batch size to 16 on a single 3090 GPU. When using 4D swin-transformer ~\cite{hong2022cost}, the batch size is 4, significantly increasing  training time.

\textbf{Ablation on support information.}
Within the domain of FSS, support sets may involve diverse data types and characteristics, which result in several patterns as shown in~ Fig.~\ref{fig:Figure1}.
Nevertheless, there is currently a lacuna in the comprehensive comparative analysis of these patterns, often attribute to a bias towards certain forms in the design of existing methods.
Considering the versatility of our framework, we evaluate all feasible combinations of guidance in Tab.~\ref{tab:input_compare}.
The Tab.~\ref{tab:input_compare} illustrates the consistent contribution of the guidance based on the class text to the performance across various scenarios.
Furthermore, the box-based settings demonstrate remarkable efficacy, notably in the class-aware box FSS, where its results eclipse those of mask FSS.
And, its performance is only marginally lower than that of class-aware mask FSS with a more expensive annotation overhead.
This presents an appealing alternative annotation paradigm for the practical application of FSS.
In Fig.~\ref{fig:pattern}, we qualitatively demonstrate the impact of different guidance types on the prediction results of query images. It can be observed that the most accurate predictions are obtained when mask, image, and text are used as guidance types of support set.

\textbf{Discussion on text guidance.}
Text, serving as a guiding cue, activates the vision-language alignment capabilities of CLIP, thereby enhancing performance. 
As shown in Tab.~\ref{tab:pascal_sota}, class-aware mask FSS achieves 8.2\% relative mIoU improvement compared to mask FSS under the 1-shot setting. 
However, increasing the number of support images has a relatively small impact on the performance improvement of class-aware mask FSS. This indicates that the text has already sufficiently activated the target in the query image. 
The categories in PASCAL-5$^i$ and COCO-20$^i$ are mainly common categories that CLIP aligns well with. Introducing text as guidance leads to performance improvement. However, introducing text on FSS-1000, which has a large number of categories, does not lead to improvement. This is because CLIP cannot align with all the categories included in FSS-1000. This indicates that introducing text does not necessarily guarantee a performance improvement. 
The insight from the above observations is that, in practical applications, text should be selectively introduced in combination with the knowledge of the pre-trained model.

\textbf{Analysis of typical failure cases.} 
We present some typical failure cases for each task pattern in Fig.~\ref{fig:fail_case}. Since class-aware mask FSS and class-aware box FSS share some common characteristics with other task patterns, we omit the failure cases for these two task patterns. 
Image FSS uses only the image as guidance, and when multiple categories appear in both the support and query images, all these categories will be activated.
Mask FSS provides additional mask annotations as guidance. These mask annotations offer rich semantic information. However, they can lead to false predictions when there is a significant variation in target scale, such as when the target is large in the support image but small in the query image.
The failure cases of box FSS are mainly due to background information within the box.
Class-aware mask FSS provides additional textual information and is the task pattern with the most guidance types. However, the rich guidance types still cannot resolve some cases of semantic confusion and scene ambiguity.
In text FSS, some similar categories are incorrectly predicted. Additionally, without any appearance information, the model cannot correctly predict camouflaged appearances based solely on text.
These problems are still challenging now. 

\subsection{Broader Impact and Limitations.}

In this paper, we attempt to consolidate the FSS tasks, which have been directed towards different research branches, into a unified technical approach.
To achieve this goal, we integrate various guidance types into a universal architecture. This may inspire the development of more general few-shot segmentation and prompt segmentation models. Although UniFSS achieves state-of-the-art performance across 7 tasks patterns, it is not yet capable of segmenting with prompts from query images in the form of points, lines, and other structures like SAM~\cite{kirillov2023segment}.


\section{Conclusion}
\label{sec:con}

In this paper, we rethink few-shot segmentation by exploring the effectiveness of multiple granular forms of guidance information in participating in query-support matching. Seven modes, including image, mask, box, text, class-aware image, class-aware mask, and class-aware box, are proposed for subsequent studies focusing on specific guidance types.
A universal vision-language baseline (UniFSS) is designed to support the fusion of intra-modal and inter-modal cues. This allows users to freely choose the guidance types and combinations for the support set beyond the single dense annotation choice of mask.
Extensive experiments demonstrate the superior performance of the proposed UniFSS on three challenging benchmarks. 
Our work will advance the development of the FSS community and promote the unification of FSS tasks.

\ifCLASSOPTIONcaptionsoff
  \newpage
\fi
\bibliographystyle{IEEEtran}
\bibliography{IEEEabrv,refs}

\begin{thebibliography}{10}
\providecommand{\url}[1]{#1}
\csname url@samestyle\endcsname
\providecommand{\newblock}{\relax}
\providecommand{\bibinfo}[2]{#2}
\providecommand{\BIBentrySTDinterwordspacing}{\spaceskip=0pt\relax}
\providecommand{\BIBentryALTinterwordstretchfactor}{4}
\providecommand{\BIBentryALTinterwordspacing}{\spaceskip=\fontdimen2\font plus
\BIBentryALTinterwordstretchfactor\fontdimen3\font minus \fontdimen4\font\relax}
\providecommand{\BIBforeignlanguage}[2]{{%
\expandafter\ifx\csname l@#1\endcsname\relax
\typeout{** WARNING: IEEEtran.bst: No hyphenation pattern has been}%
\typeout{** loaded for the language `#1'. Using the pattern for}%
\typeout{** the default language instead.}%
\else
\language=\csname l@#1\endcsname
\fi
#2}}
\providecommand{\BIBdecl}{\relax}
\BIBdecl

\bibitem{cordts2016cityscapes}
M.~Cordts, M.~Omran, S.~Ramos, T.~Rehfeld, M.~Enzweiler, R.~Benenson, U.~Franke, S.~Roth, and B.~Schiele, ``The cityscapes dataset for semantic urban scene understanding,'' in \emph{CVPR}, 2016, pp. 3213--3223.

\bibitem{sakaridis2021acdc}
C.~Sakaridis, D.~Dai, and L.~Van~Gool, ``Acdc: The adverse conditions dataset with correspondences for semantic driving scene understanding,'' in \emph{ICCV}, 2021, pp. 10\,765--10\,775.

\bibitem{min2021hypercorrelation}
J.~Min, D.~Kang, and M.~Cho, ``Hypercorrelation squeeze for few-shot segmentation,'' in \emph{ICCV}, 2021, pp. 6941--6952.

\bibitem{tian2020prior}
Z.~Tian, H.~Zhao, M.~Shu, Z.~Yang, R.~Li, and J.~Jia, ``Prior guided feature enrichment network for few-shot segmentation,'' \emph{IEEE TPAMI}, vol.~44, no.~2, pp. 1050--1065, 2020.

\bibitem{hong2022cost}
S.~Hong, S.~Cho, J.~Nam, S.~Lin, and S.~Kim, ``Cost aggregation with 4d convolutional swin transformer for few-shot segmentation,'' in \emph{ECCV}, 2022, pp. 108--126.

\bibitem{peng2023hierarchical}
B.~Peng, Z.~Tian, X.~Wu, C.~Wang, S.~Liu, J.~Su, and J.~Jia, ``Hierarchical dense correlation distillation for few-shot segmentation,'' in \emph{CVPR}, 2023, pp. 23\,641--23\,651.

\bibitem{fan2022ssp}
Q.~Fan, W.~Pei, Y.-W. Tai, and C.-K. Tang, ``Self-support few-shot semantic segmentation,'' in \emph{ECCV}, 2022.

\bibitem{shi2022dense}
X.~Shi, D.~Wei, Y.~Zhang, D.~Lu, M.~Ning, J.~Chen, K.~Ma, and Y.~Zheng, ``Dense cross-query-and-support attention weighted mask aggregation for few-shot segmentation,'' in \emph{ECCV}, 2022, pp. 151--168.

\bibitem{shuai2023pgmanet}
S.~Chen, F.~Meng, R.~Zhang, H.~Qiu, H.~Li, Q.~Wu, and L.~Xu, ``Visual and textual prior guided mask assemble for few-shot segmentation and beyond,'' \emph{IEEE TMM}, 2024.

\bibitem{liu2023delving}
X.~Liu, B.~Tian, Z.~Wang, R.~Wang, K.~Sheng, B.~Zhang, H.~Zhao, and G.~Zhou, ``Delving into shape-aware zero-shot semantic segmentation,'' in \emph{CVPR}, 2023, pp. 2999--3009.

\bibitem{shaban2017one}
A.~Shaban, S.~Bansal, Z.~Liu, I.~Essa, and B.~Boots, ``One-shot learning for semantic segmentation,'' in \emph{BMVC}, 2017, pp. 167.1--167.13.

\bibitem{moon2023msi}
S.~Moon, S.~S. Sohn, H.~Zhou, S.~Yoon, V.~Pavlovic, M.~H. Khan, and M.~Kapadia, ``Msi: maximize support-set information for few-shot segmentation,'' in \emph{ICCV}, 2023, pp. 19\,266--19\,276.

\bibitem{yang2021mining}
L.~Yang, W.~Zhuo, L.~Qi, Y.~Shi, and Y.~Gao, ``Mining latent classes for few-shot segmentation,'' in \emph{ICCV}, 2021.

\bibitem{zhang2021prototypical}
H.~Zhang and H.~Ding, ``Prototypical matching and open set rejection for zero-shot semantic segmentation,'' in \emph{ICCV}, 2021.

\bibitem{li2021adaptive}
G.~Li, V.~Jampani, L.~Sevilla-Lara, D.~Sun, J.~Kim, and J.~Kim, ``Adaptive prototype learning and allocation for few-shot segmentation,'' in \emph{CVPR}, 2021, pp. 8334--8343.

\bibitem{zhang2022mask}
G.~Zhang, S.~Navasardyan, L.~Chen, Y.~Zhao, Y.~Wei, H.~Shi \emph{et~al.}, ``Mask matching transformer for few-shot segmentation,'' in \emph{NeurIPS}, 2022, pp. 823--836.

\bibitem{zhang2021self}
B.~Zhang, J.~Xiao, and T.~Qin, ``Self-guided and cross-guided learning for few-shot segmentation,'' in \emph{CVPR}, 2021, pp. 8312--8321.

\bibitem{liu2022dynamic}
J.~Liu, Y.~Bao, G.-S. Xie, H.~Xiong, J.-J. Sonke, and E.~Gavves, ``Dynamic prototype convolution network for few-shot semantic segmentation,'' in \emph{CVPR}, 2022, pp. 11\,553--11\,562.

\bibitem{liu2022learning}
Y.~Liu, N.~Liu, Q.~Cao, X.~Yao, J.~Han, and L.~Shao, ``Learning non-target knowledge for few-shot semantic segmentation,'' in \emph{CVPR}, 2022, pp. 11\,573--11\,582.

\bibitem{zhang2022feature}
J.-W. Zhang, Y.~Sun, Y.~Yang, and W.~Chen, ``Feature-proxy transformer for few-shot segmentation,'' in \emph{NeurIPS}, 2022, pp. 6575--6588.

\bibitem{okazawa2022interclass}
A.~Okazawa, ``Interclass prototype relation for few-shot segmentation,'' in \emph{ECCV}, 2022, pp. 362--378.

\bibitem{xiong2022doubly}
Z.~Xiong, H.~Li, and X.~X. Zhu, ``Doubly deformable aggregation of covariance matrices for few-shot segmentation,'' in \emph{ECCV}, 2022, pp. 133--150.

\bibitem{wang2023focus}
Y.~Wang, N.~Luo, and T.~Zhang, ``Focus on query: Adversarial mining transformer for few-shot segmentation,'' in \emph{NeurIPS}, 2023.

\bibitem{yang2023mianet}
Y.~Yang, Q.~Chen, Y.~Feng, and T.~Huang, ``Mianet: Aggregating unbiased instance and general information for few-shot semantic segmentation,'' in \emph{CVPR}, 2023, pp. 7131--7140.

\bibitem{radford2021clip}
A.~Radford, J.~W. Kim, C.~Hallacy, A.~Ramesh, G.~Goh, S.~Agarwal, G.~Sastry, A.~Askell, P.~Mishkin, J.~Clark \emph{et~al.}, ``Learning transferable visual models from natural language supervision,'' in \emph{ICML}, 2021, pp. 8748--8763.

\bibitem{shi2022proposalclip}
H.~Shi, M.~Hayat, Y.~Wu, and J.~Cai, ``Proposalclip: Unsupervised open-category object proposal generation via exploiting clip cues,'' in \emph{CVPR}, 2022, pp. 9611--9620.

\bibitem{du2022learning}
Y.~Du, F.~Wei, Z.~Zhang, M.~Shi, Y.~Gao, and G.~Li, ``Learning to prompt for open-vocabulary object detection with vision-language model,'' in \emph{CVPR}, 2022, pp. 14\,084--14\,093.

\bibitem{zhou2022maskclip}
C.~Zhou, C.~C. Loy, and B.~Dai, ``Extract free dense labels from clip,'' in \emph{ECCV}, 2022, pp. 696--712.

\bibitem{lin2014microsoft}
T.-Y. Lin, M.~Maire, S.~Belongie, J.~Hays, P.~Perona, D.~Ramanan, P.~Doll{\'a}r, and C.~L. Zitnick, ``Microsoft coco: Common objects in context,'' in \emph{ECCV}, 2014.

\bibitem{li2020fss1000}
X.~Li, T.~Wei, Y.~P. Chen, Y.-W. Tai, and C.-K. Tang, ``Fss-1000: A 1000-class dataset for few-shot segmentation,'' in \emph{CVPR}, 2020, pp. 2869--2878.

\bibitem{dong2018few}
N.~Dong and E.~P. Xing, ``Few-shot semantic segmentation with prototype learning,'' in \emph{BMVC}, 2018.

\bibitem{snell2017prototypical}
J.~Snell, K.~Swersky, and R.~Zemel, ``Prototypical networks for few-shot learning,'' in \emph{NeurIPS}, 2017.

\bibitem{wang2019panet}
K.~Wang, J.~H. Liew, Y.~Zou, D.~Zhou, and J.~Feng, ``Panet: Few-shot image semantic segmentation with prototype alignment,'' in \emph{ICCV}, 2019.

\bibitem{vaswani2017attention}
A.~Vaswani, N.~Shazeer, N.~Parmar, J.~Uszkoreit, L.~Jones, A.~N. Gomez, {\L}.~Kaiser, and I.~Polosukhin, ``Attention is all you need,'' in \emph{NeurIPS}, 2017.

\bibitem{liu2021swin}
Z.~Liu, Y.~Lin, Y.~Cao, H.~Hu, Y.~Wei, Z.~Zhang, S.~Lin, and B.~Guo, ``Swin transformer: Hierarchical vision transformer using shifted windows,'' in \emph{ICCV}, 2021, pp. 10\,012--10\,022.

\bibitem{cho2021cats}
S.~Cho, S.~Hong, S.~Jeon, Y.~Lee, K.~Sohn, and S.~Kim, ``Cats: Cost aggregation transformers for visual correspondence,'' in \emph{NeurIPS}, 2021, pp. 9011--9023.

\bibitem{li2020correspondence}
S.~Li, K.~Han, T.~W. Costain, H.~Howard-Jenkins, and V.~Prisacariu, ``Correspondence networks with adaptive neighbourhood consensus,'' in \emph{CVPR}, 2020, pp. 10\,196--10\,205.

\bibitem{wang2023rethinking}
Y.~Wang, R.~Sun, and T.~Zhang, ``Rethinking the correlation in few-shot segmentation: A buoys view,'' in \emph{CVPR}, 2023, pp. 7183--7192.

\bibitem{siam2020weakly}
M.~Siam, N.~Doraiswamy, B.~N. Oreshkin, H.~Yao, and M.~Jagersand, ``Weakly supervised few-shot object segmentation using co-attention with visual and semantic embeddings,'' \emph{arXiv preprint arXiv:2001.09540}, 2020.

\bibitem{zhang2022weakly}
M.~Zhang, Y.~Zhou, B.~Liu, J.~Zhao, R.~Yao, Z.~Shao, and H.~Zhu, ``Weakly supervised few-shot semantic segmentation via pseudo mask enhancement and meta learning,'' \emph{IEEE TMM}, 2022.

\bibitem{wang2022imr}
H.~Wang, L.~Liu, W.~Zhang, J.~Zhang, Z.~Gan, Y.~Wang, C.~Wang, and H.~Wang, ``Iterative few-shot semantic segmentation from image label text,'' in \emph{IJCAI}, 2022, p. 1385–1392.

\bibitem{xu2021simple}
M.~Xu, Z.~Zhang, F.~Wei, Y.~Lin, Y.~Cao, H.~Hu, and X.~Bai, ``A simple baseline for zeroshot semantic segmentation with pre-trained vision-language model,'' \emph{arXiv preprint arXiv:2112.14757}, vol.~3, 2021.

\bibitem{li2022languagedriven}
\BIBentryALTinterwordspacing
B.~Li, K.~Q. Weinberger, S.~Belongie, V.~Koltun, and R.~Ranftl, ``Language-driven semantic segmentation,'' in \emph{ICLR}, 2022. [Online]. Available: \url{https://openreview.net/forum?id=RriDjddCLN}
\BIBentrySTDinterwordspacing

\bibitem{lueddecke22clipseg}
T.~L\"uddecke and A.~Ecker, ``Image segmentation using text and image prompts,'' in \emph{CVPR}, June 2022, pp. 7086--7096.

\bibitem{jia2021scaling}
C.~Jia, Y.~Yang, Y.~Xia, Y.-T. Chen, Z.~Parekh, H.~Pham, Q.~Le, Y.-H. Sung, Z.~Li, and T.~Duerig, ``Scaling up visual and vision-language representation learning with noisy text supervision,'' in \emph{ICML}, 2021, pp. 4904--4916.

\bibitem{cherti2023reproducible}
M.~Cherti, R.~Beaumont, R.~Wightman, M.~Wortsman, G.~Ilharco, C.~Gordon, C.~Schuhmann, L.~Schmidt, and J.~Jitsev, ``Reproducible scaling laws for contrastive language-image learning,'' in \emph{CVPR}, 2023, pp. 2818--2829.

\bibitem{Radford2021LearningTV}
A.~Radford, J.~W. Kim, C.~Hallacy, A.~Ramesh, G.~Goh, S.~Agarwal, G.~Sastry, A.~Askell, P.~Mishkin, J.~Clark, G.~Krueger, and I.~Sutskever, ``Learning transferable visual models from natural language supervision,'' in \emph{ICML}, 2021.

\bibitem{dou2022coarse}
Z.-Y. Dou, A.~Kamath, Z.~Gan, P.~Zhang, J.~Wang, L.~Li, Z.~Liu, C.~Liu, Y.~LeCun, N.~Peng \emph{et~al.}, ``Coarse-to-fine vision-language pre-training with fusion in the backbone,'' in \emph{NeurIPS}, 2022, pp. 32\,942--32\,956.

\bibitem{zhou2022learning}
K.~Zhou, J.~Yang, C.~C. Loy, and Z.~Liu, ``Learning to prompt for vision-language models,'' \emph{IJCV}, vol. 130, no.~9, pp. 2337--2348, 2022.

\bibitem{ding2022decoupling}
J.~Ding, N.~Xue, G.-S. Xia, and D.~Dai, ``Decoupling zero-shot semantic segmentation,'' in \emph{CVPR}, 2022, pp. 11\,583--11\,592.

\bibitem{ba2016ln}
J.~L. Ba, J.~R. Kiros, and G.~E. Hinton, ``Layer normalization,'' \emph{arXiv preprint arXiv:1607.06450}, 2016.

\bibitem{wang2023internimage}
W.~Wang, J.~Dai, Z.~Chen, Z.~Huang, Z.~Li, X.~Zhu, X.~Hu, T.~Lu, L.~Lu, H.~Li \emph{et~al.}, ``Internimage: Exploring large-scale vision foundation models with deformable convolutions,'' in \emph{CVPR}, 2023, pp. 14\,408--14\,419.

\bibitem{he2016deep}
K.~He, X.~Zhang, S.~Ren, and J.~Sun, ``Deep residual learning for image recognition,'' in \emph{CVPR}, 2016, pp. 770--778.

\bibitem{milletari2016v}
F.~Milletari, N.~Navab, and S.-A. Ahmadi, ``V-net: Fully convolutional neural networks for volumetric medical image segmentation,'' in \emph{International Conference on 3D Vision (3DV)}, 2016, pp. 565--571.

\bibitem{liu2023fecanet}
H.~Liu, P.~Peng, T.~Chen, Q.~Wang, Y.~Yao, and X.-S. Hua, ``Fecanet: Boosting few-shot semantic segmentation with feature-enhanced context-aware network,'' \emph{IEEE TMM}, 2023.

\bibitem{everingham2010pascal}
M.~Everingham, L.~Van~Gool, C.~K. Williams, J.~Winn, and A.~Zisserman, ``The pascal visual object classes (voc) challenge,'' \emph{IJCV}, vol.~88, no.~9, pp. 303--338, 2010.

\bibitem{hariharan2014simultaneous}
B.~Hariharan, P.~Arbel{\'a}ez, R.~Girshick, and J.~Malik, ``Simultaneous detection and segmentation,'' in \emph{ECCV}, 2014.

\bibitem{yao2021sdm}
X.~Yao, Q.~Cao, X.~Feng, G.~Cheng, and J.~Han, ``Scale-aware detailed matching for few-shot aerial image semantic segmentation,'' \emph{IEEE Transactions on Geoscience and Remote Sensing}, vol.~60, pp. 1--11, 2021.

\bibitem{waqas2019isaid}
S.~Waqas~Zamir, A.~Arora, A.~Gupta, S.~Khan, G.~Sun, F.~Shahbaz~Khan, F.~Zhu, L.~Shao, G.-S. Xia, and X.~Bai, ``isaid: A large-scale dataset for instance segmentation in aerial images,'' in \emph{CVPRW}, 2019, pp. 28--37.

\bibitem{paszke2019pytorch}
A.~Paszke, S.~Gross, F.~Massa, A.~Lerer, J.~Bradbury, G.~Chanan, T.~Killeen, Z.~Lin, N.~Gimelshein, L.~Antiga \emph{et~al.}, ``Pytorch: An imperative style, high-performance deep learning library,'' in \emph{NeurIPS}, 2019.

\bibitem{AdamW}
D.~P. Kingma and J.~Ba, ``Adam: A method for stochastic optimization,'' \emph{arXiv preprint arXiv:1412.6980}, 2014.

\bibitem{dosovitskiy2020image}
A.~Dosovitskiy, L.~Beyer, A.~Kolesnikov, D.~Weissenborn, X.~Zhai, T.~Unterthiner, M.~Dehghani, M.~Minderer, G.~Heigold, S.~Gelly \emph{et~al.}, ``An image is worth 16x16 words: Transformers for image recognition at scale,'' \emph{arXiv preprint arXiv:2010.11929}, 2020.

\bibitem{xian2019semantic}
Y.~Xian, S.~Choudhury, Y.~He, B.~Schiele, and Z.~Akata, ``Semantic projection network for zero-and few-label semantic segmentation,'' in \emph{CVPR}, 2019, pp. 8256--8265.

\bibitem{bucher2019zero}
M.~Bucher, T.-H. Vu, M.~Cord, and P.~P{\'e}rez, ``Zero-shot semantic segmentation,'' in \emph{NeurIPS}, vol.~32, 2019.

\bibitem{liu2022fsot}
W.~Liu, C.~Zhang, H.~Ding, T.-Y. Hung, and G.~Lin, ``Few-shot segmentation with optimal transport matching and message flow,'' \emph{IEEE TMM}, vol.~25, pp. 5130--5141, 2022.

\bibitem{zhang2023personalize}
R.~Zhang, Z.~Jiang, Z.~Guo, S.~Yan, J.~Pan, X.~Ma, H.~Dong, P.~Gao, and H.~Li, ``Personalize segment anything model with one shot,'' \emph{arXiv preprint arXiv:2305.03048}, 2023.

\bibitem{liu2023matcher}
Y.~Liu, M.~Zhu, H.~Li, H.~Chen, X.~Wang, and C.~Shen, ``Matcher: Segment anything with one shot using all-purpose feature matching,'' \emph{arXiv preprint arXiv:2305.13310}, 2023.

\bibitem{wang2023sccnet}
L.~Wang, S.~Lei, J.~He, S.~Wang, M.~Zhang, and C.-T. Lu, ``Self-correlation and cross-correlation learning for few-shot remote sensing image semantic segmentation,'' in \emph{ACM International Conference on Advances in Geographic Information Systems}, 2023, pp. 1--10.

\bibitem{kirillov2023segment}
A.~Kirillov, E.~Mintun, N.~Ravi, H.~Mao, C.~Rolland, L.~Gustafson, T.~Xiao, S.~Whitehead, A.~C. Berg, W.-Y. Lo \emph{et~al.}, ``Segment anything,'' in \emph{ICCV}, 2023, pp. 4015--4026.

\bibitem{oquab2023dinov2}
M.~Oquab, T.~Darcet, T.~Moutakanni, H.~Vo, M.~Szafraniec, V.~Khalidov, P.~Fernandez, D.~Haziza, F.~Massa, A.~El-Nouby \emph{et~al.}, ``Dinov2: Learning robust visual features without supervision,'' \emph{arXiv preprint arXiv:2304.07193}, 2023.

\end{thebibliography}
\end{document}